\documentclass[]{spieman}

\usepackage[utf8]{inputenc}
\usepackage{url}
\usepackage{graphicx}
\usepackage{amsmath}
\usepackage{amssymb}
\usepackage{booktabs}
\usepackage{longtable}
\usepackage{float}
\usepackage{xcolor}
\usepackage{tikz}
\usetikzlibrary{shapes.geometric, arrows.meta, positioning, fit, backgrounds}

\newcommand{\chg}[1]{#1}
\title{Automated Perceptually-Motivated Assessment of Photographic Consistency
in Paired Clinical Photographs: Pipeline Development and Internal Evaluation}

\author[a,*]{Derrick Lin}
\author[a]{Samantha Rabinovich}
\author[b]{Joclin Rabinovich}
\author[a]{Kassra Garoosi}
\author[a]{Sumun Khetpal}
\author[c]{Evan Delanoy}
\author[d]{Neel Bhardwaj}
\author[a]{Jason Roostaeian}

\affil[a]{Division of Plastic and Reconstructive Surgery, Department of Surgery,
David Geffen School of Medicine at UCLA, Los Angeles, California, United States}
\affil[b]{Cornell University, Ithaca, New York, United States}
\affil[c]{Tulane University, New Orleans, Louisiana, United States}
\affil[d]{University of Pittsburgh School of Medicine, Pittsburgh,
Pennsylvania, United States}

\affil[*]{Corresponding author. E-mail: DRLin@mednet.ucla.edu}

\begin{document}
\maketitle

\begin{abstract}
\textbf{Purpose:} Paired pre- and post-operative photographs are the standard
unit of evidence for plastic surgical outcomes, yet no objective metric verifies
whether two images of the same patient were captured under conditions
consistent for comparison.

\textbf{Approach:} We developed a perceptually motivated pipeline that
analyzes pre/post pairs across thirteen calibrated sub-metrics,
partitioned by unsupervised correlation-structure analysis into five data-driven
clusters (photometric, texture / sharpness, pose, illumination direction, and
\chg{pitch), averaged within each cluster and combined across clusters by
a weighted sum into a single consistency score.}
\chg{Each sub-metric is calibrated so that its median difference across
published within-patient pairs scores $0.5$, which is a reference
point and carries no pass/fail meaning.} \chg{The pipeline was calibrated on 134 matched
within-patient published pre/post pairs and evaluated against
identical-image pairs, synthetic-perturbation pairs, and 134
mismatched cross-publication pairs.}

\textbf{Results:} The master consistency
score $S$ separated matched from mismatched pairs (sensitivity index
$d' = 2.15$, 95\% confidence interval (CI) $[1.83, 2.55]$; area under
the receiver operating characteristic curve
$\mathrm{AUC} = 0.928$, 95\% CI $[0.896, 0.959]$),
closely matching Gaussian-equal-variance predictions. \chg{The three
head-pose angles did not fall in one cluster: yaw and roll grouped
together while pitch separated.} Identical pairs
scored at ceiling ($S \approx 0.99$) \chg{and the master score fell
monotonically with perturbation magnitude on all five perturbation
axes.}

\textbf{Conclusions:} The score quantifies photographic comparability, not
aesthetic or surgical quality, and \chg{provides a freely available web tool for
auditing the photographic comparability of pre/post pairs, pending
validation against expert judgment.}
\end{abstract}


% =============================================================================
% Body sections (class-agnostic fragments)
% =============================================================================
% =============================================================================
% Introduction
% =============================================================================

\section{Introduction}\label{sec:introduction}

Paired pre- and post-operative photographs are the standard visual
record in facial plastic and aesthetic surgery. Chart review,
publication, training, and medicolegal documentation often reference
them.
Each of these uses assumes the visible differences came from the
procedure. Standardized
technique exists to make that comparison
valid~\cite{DiBernardo1998}. Capture conditions
like lighting, exposure, focus, and pose vary between sessions and
can mimic or obscure real change. When that
assumption fails, the photograph shows a difference the procedure did
not cause, and every later use of the pair rests on a premise the
images cannot support.

Photographic standards for plastic surgery have existed for
nearly three decades, beginning with the DiBernardo et
al.~1998 standard \cite{DiBernardo1998} and refined in the
digital era by Prantl and colleagues \cite{Prantl2017} and at
the society level by the ASPS Photographic Guide
\cite{ASPS_Guidelines}. All three are purely \emph{prescriptive}. They
specify how to capture a photograph but offer no way to confirm
whether a given pair actually met the standard. \chg{No published plastic-surgery standard supplies an image-derived
check of its own requirements. Where conformance has been assessed, it
has been by human raters applying a
checklist~\cite{Wolfe2020}.} Consistency is left to the eye of whoever views
the pair.

Automated approaches to image quality and to objective outcome
assessment already exist. \chg{Generic image-quality metrics rate one image at a
time~\cite{Wang2004}. Face-biometrics programs compare two images to
decide whether they show the same person, and publish tolerances for
how far pose may drift between them~\cite{NIST_FRVT_2024}.
Sub-specialty outcome scores rate the surgical
result~\cite{Kim2006, Daiem2025}. None asks whether two clinical
photographs were captured alike.} They assume the two photographs were
comparable in the first place, but never check.

\chg{This paper supplies the missing check.} The pipeline reduces each pair
to 13 measurements, which group into
\chg{five interpretable clusters}, and combines them into a
single consistency score. We calibrate and evaluate it on four
datasets spanning identical self-pairs to unrelated-patient
negatives, then release it as a public web tool meant to
support the existing prescriptive standards~\cite{DiBernardo1998,
Prantl2017, ASPS_Guidelines}.

% Technical spine (Paper 1 = CS method paper). The surgeon-facing rebuild
% (methods-new/results-new) was moved to the outcome paper (Paper 2) on 2026-06-07.
% =============================================================================
%  Methods, Path A pilot manuscript
%
%  Co-write log: gutted 2026-05-17 to rebuild section by section with Derrick.
% =============================================================================

\section{Methods}

\subsection{Study Design}\label{sec:study-design}

This was a retrospective methods-development study. We built an
automated measure of photographic consistency for paired clinical
photographs and characterized how it behaves on controlled comparison
sets. No human raters were involved.

\subsection{Dataset Assembly}

All photographs analyzed in this study were obtained exclusively
from previously published peer-reviewed literature; no new patient
photographs were collected, no individual patient records were
accessed, and no patient was contacted. Under the U.S. Common Rule
definition (45~CFR~46.102(e)), the activity is not human-subjects
research, and no institutional review board review or exemption
letter was obtained. Ethics oversight for the original image
acquisition rests with the source institutions.

\subsubsection{Source photographs}
The photograph pool was assembled from a structured PubMed Boolean
search covering 2010--2025, run on 1 August 2025, combining
facial cosmetic-procedure,
photographic-documentation, and methodology terms. The full search
string is given in Supplementary Methods~\ref{sec:supp-s1}. Records
were screened independently by two reviewers (D.L., J.R.) against
the inclusion criteria in Section~\ref{sec:inclusion}, with
discrepancies resolved
by discussion to consensus. Facial photographs from retained
publications were extracted and role-tagged as pre- or
post-operative. Search yield, retention, and per-stage attrition are
reported in Section~\ref{sec:results-corpus}.

\subsubsection{Inclusion criteria and quality gates}\label{sec:inclusion}
Inclusion criteria were applied at two levels: a paper-level review of
search records against publication-type criteria, and a photo-level
review of extracted images against a sequence of automated quality
gates. At the paper level, records were reviewed against four
inclusion criteria: publication date between 2010 and 2025, publication in
a US-based journal or availability of an English translation,
retrievable patient photographs, and content depicting human facial
anatomy. Duplicate and off-topic records were removed at the same
stage.

At the photo level, extracted photographs were screened through six
sequential automated gates assessing minimum resolution, color content,
exposure, absence of text overlays, face detectability, and head-pose
recoverability. Gates were applied in the order listed, with first-fail
attribution. A photograph that failed multiple criteria was counted
only under the first failing gate. Verbatim threshold values and the
underlying detection libraries are reported in Supplementary
Methods~\ref{sec:supp-s2}. Per-stage retention and exclusion attribution are reported
in Section~\ref{sec:results-corpus} (Figure~\ref{fig:prisma}).

\subsection{Comparison Sets}\label{sec:comparison-sets}
Eligible photographs were assembled into four comparison sets, each
serving a different role: trivial positives (Set~1), a sensitivity
sweep (Set~2), real pre/post pairs (Set~3), and mismatched pairs
(Set~4). Both members of every pair had to pass all
photo-level inclusion gates. The full source-publication provenance is reported in
Supplementary Methods~\ref{sec:supp-s1}. Per-set pair counts, unique-photograph
counts, contributing-publication counts, and composition are reported
in Section~\ref{sec:results-set-assembly}
(Table~\ref{tab:set-composition}).

\paragraph{Set 1, Trivial positives.}
Each eligible photograph was paired with itself ($n = 309$). The two
images are identical, so these pairs should score at the top of the
range. Set~1 establishes that upper bound and confirms the pipeline
reaches it. Failure on this set indicates instability in the
measurement pipeline itself.

\paragraph{Set 2, Sensitivity sweep.}
Twelve source photographs drawn at random from the eligible pool were
perturbed along five axes, one axis at a time: Gaussian blur, global brightness,
white-balance shift, in-plane rotation, and uniform scale. Across 43
magnitude levels in total this yielded $12 \times 43 = 516$
source-vs-perturbed pairs and a magnitude--response curve per axis.
Metrics governed by three-dimensional capture geometry (out-of-plane
pose, illumination direction; Section~\ref{sec:metrics}) cannot be
synthesized as controlled 2D
perturbations, so Set~2 covers the synthesizable axes and leaves the
rest to Set~3. Controlled-distortion
testing of this kind is standard in the assessment of image-quality
metrics~\cite{Wang2004}. Set~2 is not used to fit the cluster weights
(Section~\ref{sec:cluster-weights}). Per-axis
units and magnitude values are reported in Supplementary
Methods~\ref{sec:supp-s1}.

\paragraph{Set 3, Real pre/post pairs.}
Within-patient pre- and post-operative photographs from the same
publication were paired by patient identifier as given in the source
publication ($n = 134$). These are real clinical photographs taken
across an operative interval, under the lighting and positioning
variation typical of routine before-and-after documentation. Set~3 is
the reference set against which all sub-metrics are calibrated
(Section~\ref{sec:metrics-jnd}). It is also the positive class in the
reported discriminant, so the same pairs both calibrate and evaluate
the score. Section~\ref{sec:cluster-weights} bounds the resulting
optimism.

\paragraph{Set 4, Mismatched pairs.}
Cross-study pairs were constructed by randomly pairing one patient's
photograph with an unrelated patient's photograph from a different
publication ($n = 134$), equal in count to Set~3. The two images come
from independent clinical settings, equipment, and subjects, so these
are inconsistent pairs by construction, and they are the negative class
in the Set~3-versus-Set~4 discriminant.

\subsection{Photographic Consistency Metrics}\label{sec:metrics}
Photographic consistency between a pair of images is summarized by a
single overall score $S$ on $[0, 1]$, with 1 indicating maximum
agreement and 0 indicating maximum disagreement.

The score combines 13 sub-metrics: brightness, contrast, shadow
extent, global and median-pixel $\Delta E_{00}$, CIELAB lightness
offset, image gradient, focus, noise, yaw, pitch, roll, and
illumination direction. The sub-metrics are not grouped in advance.
They are partitioned into clusters by an unsupervised correlation
analysis.

Each sub-metric measures a capture condition that published
\chg{photographic standards require to be held constant: illumination,
film and sensor response, and patient
position}~\cite{DiBernardo1998, Prantl2017, ASPS_Guidelines}. A
systematic review of facial photography adds exposure, focus, cast
shadow, and view~\cite{Wolfe2020}. A
sub-metric was included when its requirement could be recovered from
the image pair alone, without capture metadata. \chg{Requirements with no
corresponding sub-metric are either unrecoverable from the pair or
were not implemented}
(Section~\ref{sec:discussion-limitations}). Noise is the one exception
to that rule, included on image-quality grounds~\cite{Wang2004} because
no published standard specifies a noise requirement. The full
requirement-to-sub-metric correspondence is given in Supplementary
Methods~\ref{sec:supp-provenance}.

A calibrated score of $0.5$, on a sub-metric or on the master score
$S$, marks the median disagreement in routine published pre/post pairs
(Set~3). It carries no pass/fail or perfect-to-chance meaning.

The remainder of this subsection describes how each sub-metric is
calibrated to a common $[0, 1]$ scale, including the full
empirical-anchor argument (Section~\ref{sec:metrics-jnd}), and how
calibrated sub-metric scores are aggregated into the overall score
(Sections~\ref{sec:metrics-agg}--\ref{sec:cluster-weights}).

\subsubsection{Empirical calibration to a common \texorpdfstring{$[0, 1]$}{[0,1]} scale}\label{sec:metrics-jnd}
All sub-metrics are calibrated to Set~3, because it is the only set
with a spread of differences between pre/post photographs. Each
calibration is fit so that the median Set~3 raw difference maps to a
calibrated score of $0.5$. Supplementary
Methods~\ref{sec:supp-s6} tests an alternative anchoring at the
endpoints, mapping Set~1 to 1 and the 95th percentile of Set~4 raw
differences to 0, and reports how far the headline results move.

Each sub-metric is converted into a $[0, 1]$ consistency score in two
steps. First, the raw absolute difference between the two images is
passed through a metric-specific monotone transform (square-root,
$\ln(1+x)$, or identity) chosen to stabilize variance. Brightness and
lightness follow CIELAB $L^{*}$~\cite{CIE15_2004,Mahy1994}, color
difference follows CIEDE2000~\cite{Sharma2005}, and the rotational
angles follow operational head-pose
tolerances~\cite{NIST_FRVT_2024,Wilson2000}.

Second, the transformed difference is mapped to a score by one of two
backends. A Signal Detection Theory (SDT)~\cite{GreenSwets1966} mapping
(Equation~\ref{eq:s3-sdt}) serves the five sub-metrics whose transformed
Set~3 distributions are well-approximated by a Gaussian (brightness,
contrast, global and median-pixel $\Delta E_{00}$, and the CIELAB
lightness offset). A
direct empirical-CDF lookup against the Set~3 distribution
(Equation~\ref{eq:s3-ecdf}) serves the remaining eight, whose raw
differences are bounded, heavy-tailed, or zero-inflated. The
assignment is fixed in advance.

The Signal Detection Theory backend implements the median anchor
through a fitted Just-Noticeable-Difference ($\mathrm{JND}_t$, in
transformed units). The JND is therefore
a calibration parameter, despite the perceptual name. Literature values
such as NIST FRVT pose tolerances and CIEDE2000 $\Delta E$ serve as
cross-checks on the fitted value. Per-metric transforms, backend assignments, fitted JND
values, and the underlying perceptual references are reported in
Supplementary Methods~\ref{sec:supp-s3}. An empirical walk-through of
the procedure on the global $\Delta E_{00}$ sub-metric is presented in
Section~\ref{sec:results-calibration}
(Figure~\ref{fig:calibration-walkthrough}).

\subsubsection{Aggregation and cluster weights}\label{sec:metrics-agg}
Calibrated sub-metric scores are aggregated into the overall
photographic consistency score in two stages. First, the thirteen
sub-metrics are partitioned into clusters on Set~3 alone, grouping
sub-metrics whose calibrated scores co-vary across real within-patient
pre/post pairs. The partition uses average-linkage hierarchical
clustering on a $1-|\rho|$ distance, where $\rho$ is the Pearson
correlation between sub-metric calibrated-score vectors across Set~3
pairs, cut at height $0.75$ (Supplementary
Methods~\ref{sec:supp-s4}). Each
cluster score is the unweighted mean of its member calibrated scores,
and the discovered cluster count and membership, with the interpretive
labels, are reported in Section~\ref{sec:results-cluster-structure}.
Clustering on Set~2 instead would encode the experimenter-chosen
perturbation axes.

Second, the cluster scores $\bar{s}_c$ are combined into the master
consistency score by a weighted sum,
\begin{equation}
S \;=\; \sum_{c} w_{c}\,\bar{s}_{c},
\qquad \sum_{c} w_{c} = 1,
\label{eq:master}
\end{equation}
where $w_c$ is the weight of cluster $c$. The cluster weights are fit by the
discriminant-maximization procedure described in the next section, and
the fitted values are reported in
Section~\ref{sec:results-weight-fit}. Clustering limits the
contribution of correlated sub-metrics, which would otherwise enter the
weighted sum as near-redundant terms. Clusters are also the unit at
which discriminant performance is reported
(Section~\ref{sec:results-discriminant}).

\subsection{Cluster weight fitting: discriminant maximization}\label{sec:cluster-weights}
Clusters differ in how much they separate Set~3 from Set~4.
Calibration equalizes each sub-metric's median and leaves its
discriminative power untouched.

The cluster weights are fit by directly maximizing the between-set
discriminant, the sensitivity index $d'$ between the master scores on
Set~3 (real within-patient pre/post) and Set~4 (mismatched pairs),
\begin{equation}
d' \;=\; \frac{\mu_{S_3} - \mu_{S_4}}{\sigma_{\text{pooled}}},
\qquad
\sigma_{\text{pooled}}^{2} \;=\; \tfrac{1}{2}\bigl(\sigma_{S_3}^{2} + \sigma_{S_4}^{2}\bigr),
\label{eq:dprime}
\end{equation}
following the standard SDT
definition~\cite{GreenSwets1966,MacmillanCreelman2005}, where $\mu$ and
$\sigma$ are the mean and standard deviation of the
master score on the indicated set. The fit objective is the same
statistic the master is evaluated on (Section~\ref{sec:eval}), so the
headline discriminant is not held out in the strict sense. No
human-rated consistency labels exist, and fitting instead on Set~2
would tie the weights to the perturbation design.

They are constrained to be non-negative, to sum to one, and to lie
within $[0.02, 0.70]$, and are obtained by sequential
least-squares quadratic programming (SLSQP) from $24$ restarts, one
uniform start plus $23$ symmetric-Dirichlet draws, with random seed
$0$. All $24$ restarts converged.

Cross-validation re-estimates the cluster weights on each of five
folds (seed $0$), with the sub-metric calibration and the cluster
partition held fixed at their full-Set~3 values. This bounds optimism
from the weight fit. It does not bound optimism from the calibration or
the clustering, which are unsupervised with respect to the
Set~3-versus-Set~4 contrast. A group split by source publication
(Supplementary Methods~\ref{sec:supp-s4}) additionally guards against
leakage between folds.

\subsection{Discriminant evaluation}\label{sec:eval}
Separation between two comparison sets is reported as the sensitivity
index $d'$~\cite{GreenSwets1966,MacmillanCreelman2005} of
Equation~\ref{eq:dprime}. For interpretability, $d'$ is also mapped to
the area under the receiver operating characteristic curve (AUC)
under an equal-variance Gaussian model~\cite{MacmillanCreelman2005},
\begin{equation}
\mathrm{AUC} \;=\; \Phi\!\bigl(d'/\sqrt{2}\bigr),
\label{eq:auc}
\end{equation}
where $\Phi$ is the standard normal cumulative distribution function.
The AUC computed directly from the score distributions is reported
alongside it as the empirical AUC.

Set~3 versus Set~4 is the primary development discriminant, and is
recomputed on each cluster score $\bar{s}_c$ to give a per-cluster $d'$
(Section~\ref{sec:results-discriminant}). Set~1 and Set~2 are reported as
mean scores. Set~1 has zero within-set variance. Any $d'$ computed
against it rescales a mean difference.

\paragraph{Bootstrap confidence intervals.} $d'$, empirical AUC, mean scores, and the Set~3--Set~4 mean-score difference are reported with 95\% bootstrap percentile confidence intervals. Resampling draws whole source publications, because images from one publication share capture conditions and resampling individual pairs would underestimate variance.

Each iteration draws 51 source publications at random with replacement, the number contributing to Set~3 or Set~4, and rebuilds both sets from those drawn. A Set~3 pair enters once for each time its publication is drawn. A Set~4 pair spans two publications, so it enters once if both are drawn and is not duplicated if either is drawn twice. The statistics are recomputed on the rebuilt sets, once per iteration. The 95\% confidence interval for each runs from the 2.5th to the 97.5th percentile of its $B = 5{,}000$ values. The same procedure is applied per cluster, and the random seed and bootstrap output are released with the code (Section~\ref{sec:implementation}).

\subsection{Within-cohort grading of clinical pre/post pairs}\label{sec:within-cohort-eval}
Set~3 is split into tertiles by master score, and the mean
per-cluster score is computed in each tertile. This tests whether the
master score separates pairs within a single clinical cohort as well
as between sets. The top-minus-bottom-tertile difference
$\Delta_c = \bar{s}_c^{\text{top}} - \bar{s}_c^{\text{bottom}}$
identifies which clusters carry the within-cohort signal.

\subsection{Implementation: web-based analysis tool}\label{sec:implementation}
The full pipeline of Sections~\ref{sec:metrics}--\ref{sec:within-cohort-eval}
is packaged as a self-contained web application to make the
consistency analysis usable at the point of care without local
installation. The analysis backend wraps the pipeline of
Section~\ref{sec:metrics} in a Python REST API (FastAPI, Uvicorn)
exposing single-pair and batch endpoints; batch jobs stream
pair-level progress to the client over WebSockets. The frontend is
a single-page React/TypeScript client (Vite, Tailwind) that uploads
image pairs, reports the master and cluster scores as percentiles
of the Set~3 distribution, and offers side-by-side, overlay, and
difference views of the pair.
Sub-metric implementations (Section~\ref{sec:metrics}), face
alignment (MediaPipe), head pose (6DRepNet), and the photometric,
color, and quality metrics, are shared between the analysis
scripts that produced the results below and the deployed tool, so
the manuscript scores and the tool scores are bit-identical for the
same input pair.

\paragraph{Image reproduction.}
Clinical photographs reproduced as figures in this manuscript and
its supplement (Figure~\ref{fig:s1-perturbation-panel};
Figure~\ref{fig:ui-screenshot}) are taken from an open-access source
publication distributed under a Creative Commons Attribution (CC~BY)
licence~\cite{khan2019satisfaction}, which permits reproduction and
modification with attribution. Both figures are attributed in their
captions, and the perturbed panels of
Figure~\ref{fig:s1-perturbation-panel} are identified there as
modifications. No clinical photographs are reproduced beyond what is
necessary to illustrate the method.

\paragraph{Web-tool data handling.}
The deployed web tool is intentionally stateless. Uploaded image
pairs are written to an ephemeral per-job temporary directory,
passed to the analysis pipeline, and deleted at the end of the
request; no image bytes are persisted to disk, no analysis results
are cached server-side, and image bytes do not appear in
application logs. The tool is hosted on a managed ephemeral
container platform whose container model additionally clears state
on container restart.

\paragraph{Intended use.}
The tool is intended for methodological and research use, specifically, for auditing the photographic comparability of pre/post
pairs prior to outcome analysis or before-and-after publication. It
is not a clinical diagnostic device, does not produce a clinical
determination, and has not been evaluated against any regulatory
framework for clinical decision support. Users are advised against
uploading identifiable patient photographs to the public deployment;
for clinical-image workflows, the implementation should be obtained
from the corresponding author and run locally.

%% =============================================================================
%%  results.tex, Path A pilot manuscript
%%  Co-write log: drafted 2026-05-17 starting with §3.1 discriminant headline.
%% =============================================================================

\section{Results}\label{sec:results}

\subsection{Photograph pool assembly}\label{sec:results-corpus}

Applying the search strategy and inclusion criteria of
Section~\ref{sec:inclusion}, the PubMed search returned 282 records.
Of these, 185 failed at least one paper-level criterion: photographs
not retrievable ($n = 127$), outside the date range ($n = 28$), non-US
journal without an English translation ($n = 21$), not depicting facial
or human subjects ($n = 7$), and duplicate or off-topic ($n = 2$). The
remaining 97 publications went to image extraction.

From those publications, 974 facial photographs were extracted and
screened through the six photo-level gates. First-fail attribution
excluded 665 photographs (68.3\%): below minimum resolution
($n = 380$), face undetected ($n = 203$), exposure problem ($n = 43$),
grayscale or low color ($n = 20$), text overlay ($n = 19$), and pose
unrecoverable ($n = 0$). This left 309 eligible photographs from 52
contributing publications (Figure~\ref{fig:prisma}). Per-publication yield is
summarized in Supplementary Methods~\ref{sec:supp-s1}.

% PRISMA figure (single source: prisma_figure.tex; \label{fig:prisma} defined there).
% PRISMA-style photograph pool assembly figure. \input{} from results.tex;
% \label{fig:prisma} is defined here (single source).
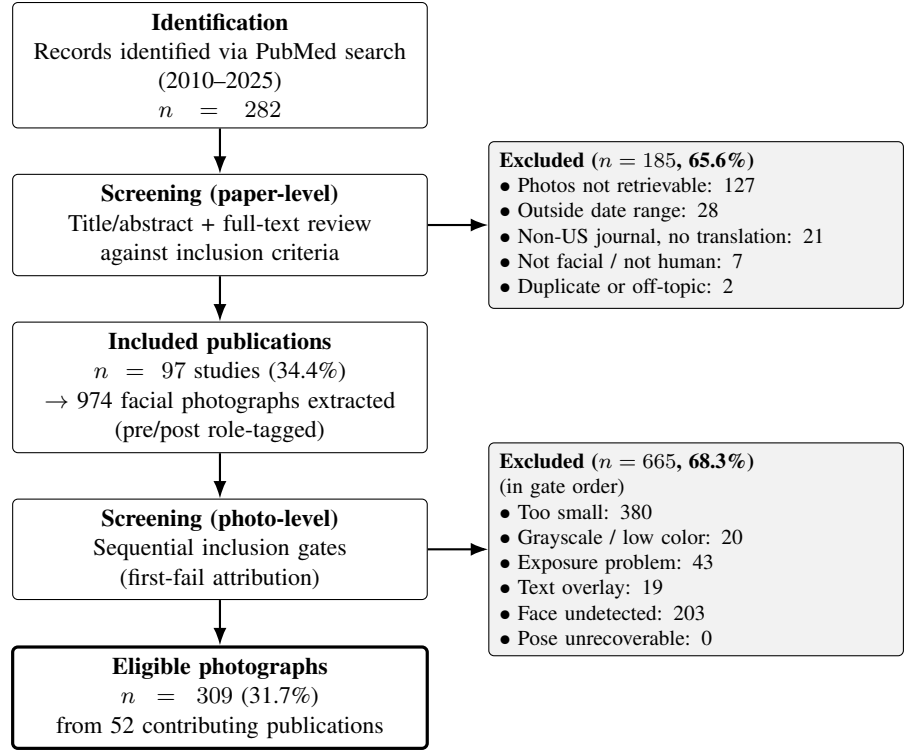
\begin{figure}[tbp]
  \centering
  \begin{tikzpicture}[
    node distance=6mm and 8mm,
    stage/.style={draw, rounded corners=2pt, align=center,
                  text width=52mm, inner sep=4pt, font=\small},
    excl/.style ={draw, rounded corners=2pt, align=left,
                  text width=52mm, inner sep=4pt, font=\footnotesize,
                  fill=gray!10},
    arr/.style  ={-Latex, thick},
  ]

    % --- Tier 1: paper-level ---
    \node[stage] (id1) {\textbf{Identification}\\Records identified via PubMed search\\(2010--2025)\\$n = 282$};
    \node[stage, below=of id1] (screen1) {\textbf{Screening (paper-level)}\\Title/abstract + full-text review\\against inclusion criteria};
    \node[excl, right=of screen1] (exclP)
      {\textbf{Excluded ($n = 185$, 65.6\%)}\\
       $\bullet$~Photos not retrievable: 127\\
       $\bullet$~Outside date range: 28\\
       $\bullet$~Non-US journal, no translation: 21\\
       $\bullet$~Not facial / not human: 7\\
       $\bullet$~Duplicate or off-topic: 2};
    \node[stage, below=of screen1] (incP) {\textbf{Included publications}\\$n = 97$ studies (34.4\%)\\$\rightarrow$ 974 facial photographs extracted\\(pre/post role-tagged)};

    % --- Tier 2: photo-level ---
    \node[stage, below=of incP] (screen2) {\textbf{Screening (photo-level)}\\Sequential inclusion gates\\(first-fail attribution)};
    \node[excl, right=of screen2] (exclI)
      {\textbf{Excluded ($n = 665$, 68.3\%)}\\
       (in gate order)\\
       $\bullet$~Too small: 380\\
       $\bullet$~Grayscale / low color: 20\\
       $\bullet$~Exposure problem: 43\\
       $\bullet$~Text overlay: 19\\
       $\bullet$~Face undetected: 203\\
       $\bullet$~Pose unrecoverable: 0};
    \node[stage, very thick, below=of screen2] (incI) {\textbf{Eligible photographs}\\$n = 309$ (31.7\%)\\from 52 contributing publications};

    % --- Arrows ---
    \draw[arr] (id1) -- (screen1);
    \draw[arr] (screen1) -- (exclP);
    \draw[arr] (screen1) -- (incP);
    \draw[arr] (incP) -- (screen2);
    \draw[arr] (screen2) -- (exclI);
    \draw[arr] (screen2) -- (incI);

  \end{tikzpicture}
  \caption{PRISMA-style photograph pool assembly flow. Paper-level screening reduced 282
  PubMed records to 97 included publications, from which 974 facial photographs
  were extracted. Photo-level screening applied six sequential inclusion gates
  (first-fail attribution) covering minimum resolution, color content, exposure,
  absence of overlaid text, face detectability, and head-pose recoverability,
  retaining 309 photographs (31.7\%) from 52 contributing publications. Assembly
  of these eligible photographs into the four comparison sets is reported in
  Section~\ref{sec:results-set-assembly} and Table~\ref{tab:set-composition}.}
  \label{fig:prisma}
\end{figure}

\subsection{Comparison-set assembly}\label{sec:results-set-assembly}

The 309 eligible photographs were assembled into the four comparison
sets defined in Section~\ref{sec:comparison-sets}, yielding a final
evaluation pool of $1{,}093$ image pairs. Per-set pair counts,
unique-photograph counts, contributing-publication counts, and
composition are reported in Table~\ref{tab:set-composition}. Set~3
(real within-patient pre/post) used 268 of the 309 eligible
photographs, drawn from 46 of the 52 contributing publications. The
remaining 41 photographs had no within-patient counterpart. Six
publications contributed none at all: their photographs passed the
photo-level gates but yielded no pre/post pair for the same patient
under the pairing rule of Section~\ref{sec:comparison-sets}. These are
single-photo case reports, cross-sectional series, or pre-only and
post-only cohorts. Set~4 (mismatched pairs) draws on 50 of the 52 contributing
publications; one publication contributes eligible photographs but
appears in neither Set~3 nor Set~4.

\begin{table}[h]
\centering
\caption{Composition of the four comparison sets. ``Unique
photographs'' counts distinct source photographs appearing in the set
(a photograph may participate in multiple pairs; for Set~2 the
perturbed variants are not counted separately). ``Source publications''
counts the number of contributing publications in the eligible pool
(52 in total). Pair-construction conventions for each set are
footnoted below.}
\label{tab:set-composition}
\begin{tabular}{@{}lrrr@{}}
\toprule
Set & Pairs & Unique photographs & Source publications \\
\midrule
1. Trivial positives\textsuperscript{a}  & 309 & 309 & 52 \\
2. Sensitivity sweep\textsuperscript{b}  & 516 &  12 & 11 \\
3. Real pre/post\textsuperscript{c}      & 134 & 268 & 46 \\
4. Mismatched pairs\textsuperscript{d}  & 134 & 179 & 50 \\
\midrule
Total & $1{,}093$ & --- & --- \\
\bottomrule
\end{tabular}

\vspace{4pt}
\begin{flushleft}\footnotesize
\textsuperscript{a}Each eligible photograph paired with itself; establishes the upper-bound identity ceiling.\\
\textsuperscript{b}12 source photographs perturbed at 43 magnitudes across 5 single-axis perturbations (Gaussian blur, brightness, white-balance, rotation, scale). 484 of the 516 constructed pairs could be scored; 32 failed face detection at the strongest perturbations.\\
\textsuperscript{c}Within-patient pre- and post-operative pairs drawn from the same publication.\\
\textsuperscript{d}Random cross-publication pairs, equal in count to Set~3; both members from different patients in different publications.
\end{flushleft}
\end{table}

\subsection{Sub-metric calibration: a worked example}\label{sec:results-calibration}

\begin{figure}[h]
\centering
\includegraphics[width=\linewidth]{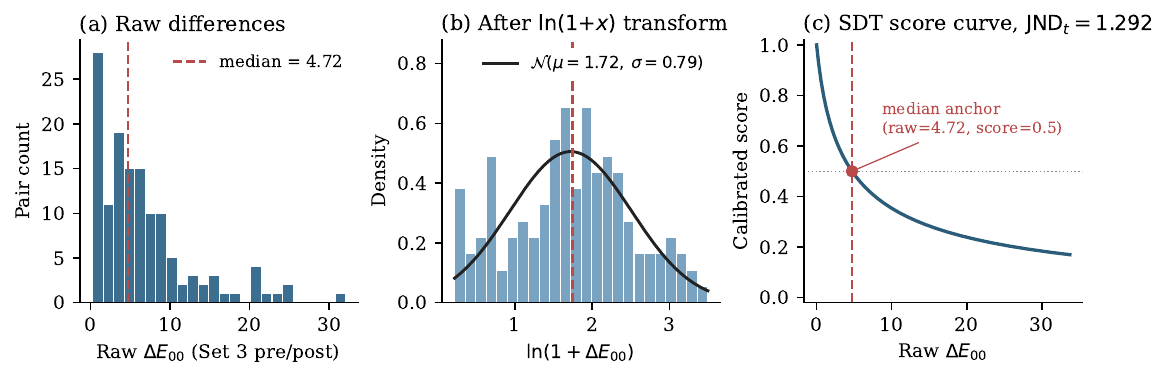}
\caption{Single-metric calibration walk-through for global
$\Delta E_{00}$ on Set~3 ($n = 134$ within-patient pre/post pairs).
(a) Distribution of raw absolute pairwise CIEDE2000 differences; the
red dashed line marks the Set~3 median. (b) Same data after the
variance-stabilizing $\ln(1{+}x)$ transform, overlaid with a Gaussian
fit ($\mu$, $\sigma$ estimated from the transformed sample) used by the
Signal Detection Theory backend. (c) Resulting calibrated score curve
(Equation~\ref{eq:s3-sdt}), with $\mathrm{JND}_t$ fit so
that the Set~3 median raw difference maps to a calibrated score of
$0.5$ (red marker).}
\label{fig:calibration-walkthrough}
\end{figure}

To make the calibration procedure of Section~\ref{sec:metrics-jnd}
concrete, Figure~\ref{fig:calibration-walkthrough} traces it end-to-end
for one sub-metric, the global CIEDE2000 color difference
($\Delta E_{00}$), on Set~3 ($n = 134$ within-patient pre/post pairs).
The raw absolute pairwise $\Delta E_{00}$ values
(Figure~\ref{fig:calibration-walkthrough}a) are right-skewed
(skewness $+1.69$). The variance-stabilizing $\ln(1{+}x)$ transform
removes that skew (skewness $-0.04$) and brings the sample close to
Gaussian (Shapiro--Wilk $W = 0.98$;
Figure~\ref{fig:calibration-walkthrough}b), as the Signal Detection
Theory backend assumes. The fitted
Just-Noticeable-Difference is $\mathrm{JND}_t = 1.292$, which maps the median Set~3 raw difference to
a calibrated score of $0.5$
(Figure~\ref{fig:calibration-walkthrough}c). Every sub-metric follows the same two steps, with its own transform
and backend. Supplementary Methods~\ref{sec:supp-s3} tabulates the
transform, backend, and fitted JND for each.

\subsection{Cluster structure}\label{sec:results-cluster-structure}

\begin{figure}[h]
\centering
\includegraphics[width=\linewidth]{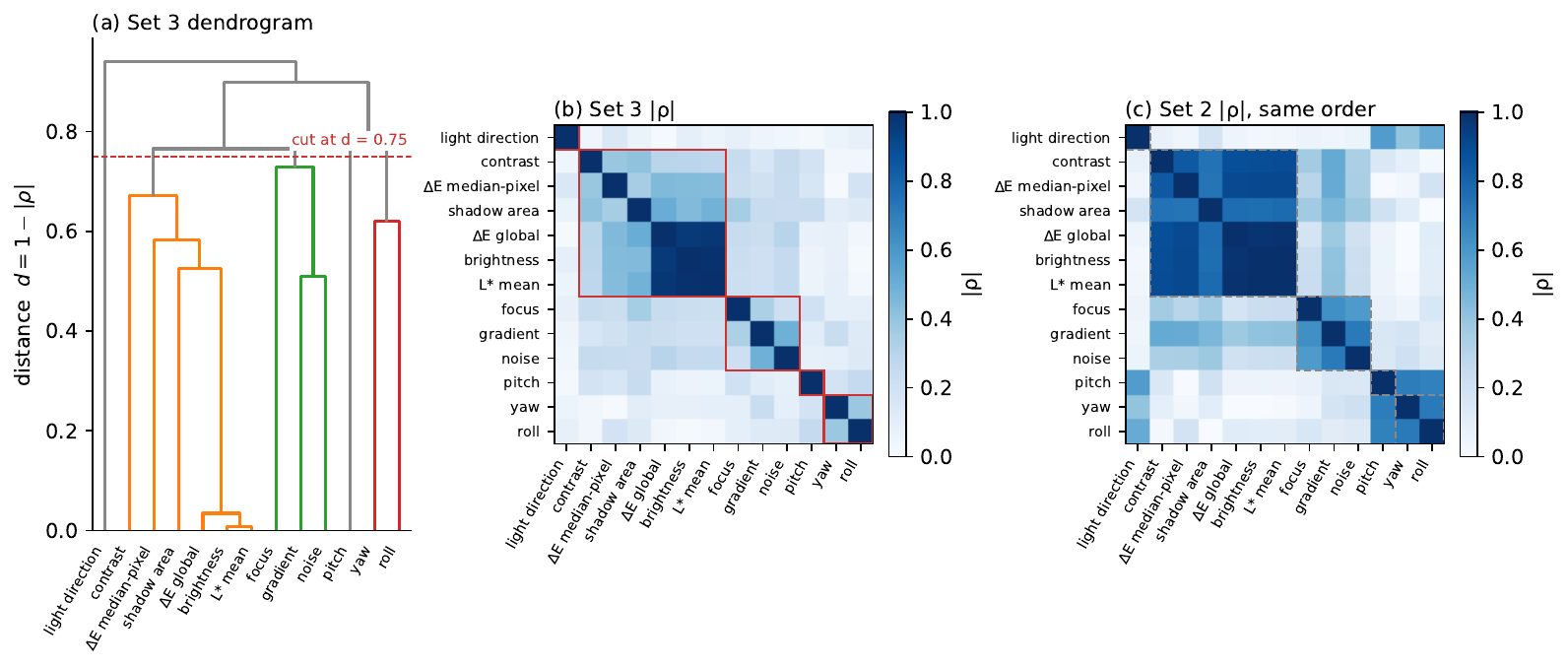}
\caption{Cluster structure of the thirteen calibrated sub-metrics,
defined on Set~3. (a)~Average-linkage dendrogram on the Set~3
correlation distance (Section~\ref{sec:metrics-agg}). The dashed red
line marks the cut at $0.75$, yielding five clusters (C1--C5) used throughout the
discriminant fit and per-cluster analyses
(Section~\ref{sec:results-discriminant}).
(b)~Set~3 Pearson $|\rho|$ matrix with rows and columns ordered by
dendrogram leaves; red rectangles outline the five clusters.
(c)~Set~2 Pearson $|\rho|$ matrix, plotted in the same leaf order
with the same cluster outlines overlaid as dashed gray. On Set~2 the
three pose metrics (pitch, yaw, roll) bundle into a single block,
whereas on Set~3 pitch (C5) separates from pose (C3), so the split is
not a clustering-protocol artifact. The cluster definition used
throughout is the one derived from all of Set~3 (Supplementary
Methods~\ref{sec:supp-s4}).}
\label{fig:cluster-structure}
\end{figure}

Average-linkage hierarchical clustering on the Set~3 correlation
distance (Section~\ref{sec:metrics-agg}) between the thirteen
calibrated sub-metrics returned five clusters at a cut of $0.75$
(Figure~\ref{fig:cluster-structure}a; Supplementary
Methods~\ref{sec:supp-s4}). The cut sits at the broadest natural gap in the
dendrogram; coarser cuts merge groups with different photographic meanings
(notably pitch with yaw and roll), while finer cuts fragment the
photometric block. The five clusters, their member sub-metrics, and their mean
within-cluster $|\rho|$ on Set~3 are:
\begin{itemize}\setlength{\itemsep}{0pt}
  \item \textbf{C1 photometric}: brightness, contrast, shadow extent,
        global and median-pixel $\Delta E_{00}$, CIELAB lightness
        offset ($|\rho| = 0.51$)
  \item \textbf{C2 texture/sharpness}: image gradient, focus, noise
        ($|\rho| = 0.34$)
  \item \textbf{C3 pose}: yaw, roll ($|\rho| = 0.38$)
  \item \textbf{C4 illumination direction}: dominant-illuminant
        angular distance (singleton)
  \item \textbf{C5 pitch}: pitch (singleton)
\end{itemize}
The procedure determined the number of clusters and their membership.
We assigned the labels afterward, by inspecting what each group has in
common photographically.

Clustering on Set~2 alone recovers a single three-member pose cluster
(Figure~\ref{fig:cluster-structure}c), because Set~2's in-plane
rotation axis perturbs only roll, leaving pitch and yaw coupled in its
correlation structure only through their shared zero-response
baseline.

\subsection{Cluster weight fit: discriminant maximization}\label{sec:results-weight-fit}

The fitted cluster weights $w_c$ (Equation~\ref{eq:master}) were
$w_1 = 0.339$ (C1 photometric),
$w_2 = 0.262$ (C2 texture/sharpness), $w_3 = 0.108$ (C3 pose), $w_4 = 0.153$ (C4 illumination direction), and $w_5 =
0.138$ (C5 pitch). No weight saturated against the bounds.

Fitting the weights raised the between-set discriminant
(Equation~\ref{eq:dprime}) from $d' = 2.03$ at equal weights to
$d' = 2.15$ on the full Set~3/Set~4 contrast. Under $5$-fold
cross-validation (Section~\ref{sec:cluster-weights}) the fitted
weights give $d' = 2.13 \pm 0.42$ (mean $\pm$ SD across folds), against
$2.15$ on the full data. Supplementary
Methods~\ref{sec:supp-s4}~(Table~\ref{tab:agg-sensitivity}) compares
the five-cluster weighted aggregation against alternative schemes.

A stricter cross-validation excludes any pair
sharing a publication with the test fold
(Supplementary Methods~\ref{sec:supp-s4},
\hyperref[sec:supp-s431]{Section~S4.3.1}). The fitted weights give
a pooled held-out $d' = 2.20$ (empirical AUC $0.936$), against an
equal-weight baseline of $2.13$ under the same splits.

\begin{figure}[h]
\centering
\includegraphics[width=\linewidth]{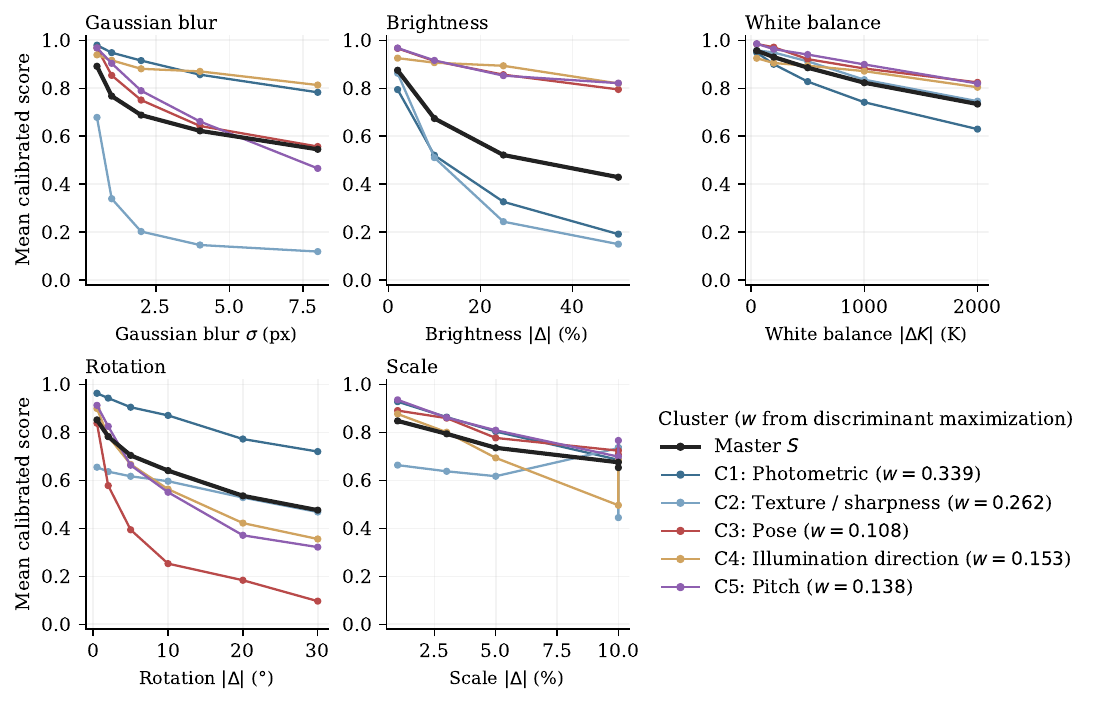}
\caption{Set~2 magnitude--response curves by cluster and master, an
independent check on the cluster aggregation.
Each panel plots the mean calibrated score (averaged across the 12
Set~2 source photographs at each magnitude level) against the
perturbation magnitude in native units for one of the five Set~2
axes (Gaussian blur $\sigma$ in px, brightness $|\Delta|$ in \%,
white-balance shift $|\Delta K|$ in K, rotation $|\Delta|$ in
degrees, scale factor). The colored lines show the five cluster scores
(C1 photometric, C2 texture/sharpness, C3 pose,
C4 illumination direction, C5 pitch); the heavy black line is the
master consistency score $S$ computed with the
discriminant-maximization cluster weights. The
master
decreases monotonically with magnitude on all five axes. Cluster-level curves show the expected
within-cluster sensitivities: texture/sharpness collapses under blur,
photometric tracks brightness and white balance, and pose responds to
in-plane rotation. No single cluster dominates the master.}
\label{fig:dose-response}
\end{figure}

Set~2 provides an independent magnitude--response check on the
aggregation. Of the 516 constructed pairs, 484 could be scored. The other 32 failed
face detection at the strongest perturbation levels. On the scored pairs the master score decreases monotonically with
perturbation magnitude on all five Set~2 axes
(Figure~\ref{fig:dose-response}). The cluster scores do the same on
four axes; on the scale axis, four of the five rise at one magnitude
step.

\subsection{Discriminant performance}\label{sec:results-discriminant}

The master consistency score $S$, computed with the cluster weights of
Section~\ref{sec:results-weight-fit}, ordered Sets~1, 3 and~4 as
their construction predicts (Table~\ref{tab:by-set}): identical
pairs at the ceiling ($S = 0.990$), real within-patient pre/post pairs
next, and mismatched pairs lowest. The highest-scoring real pair
reached $S = 0.793$, and the lowest $S = 0.183$.

\begin{figure}[h]
\centering
\includegraphics[width=\linewidth]{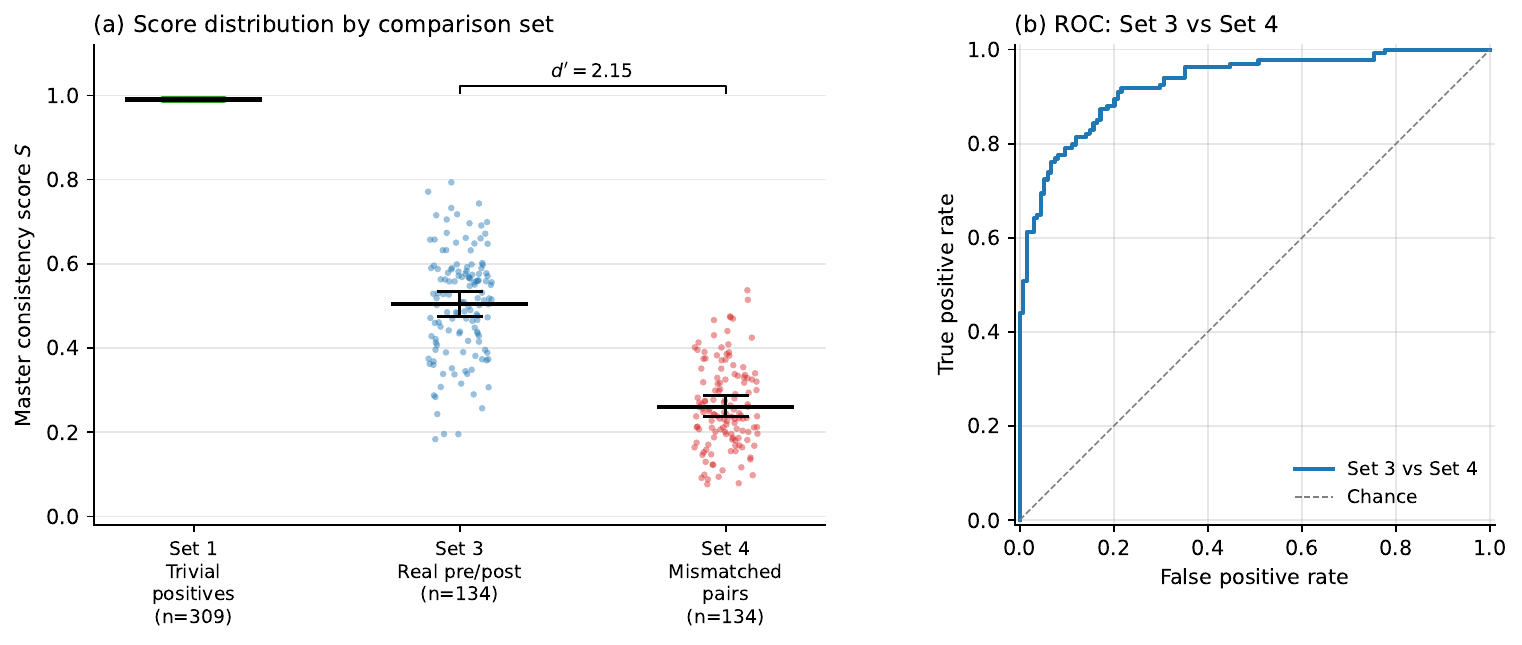}
\caption{Discriminant performance of the master consistency score $S$.
(a) Per-pair scores by comparison set (Sets~1, 3, and 4). Horizontal
bars show set means, and whiskers the 95\% bootstrap CI of the mean over
resampled source publications. Set~1 has zero within-set variance and carries
neither a whisker nor a $d'$ bracket; any $d'$ computed against it
rescales a mean difference. Set~2 (sensitivity sweep) is omitted. The bracket gives $d'$ for the primary development
contrast, Set~3 versus Set~4 (Section~\ref{sec:eval}).
(b) Receiver operating characteristic for the same contrast, with Set~3
as the positive class. The empirical area under the curve is $0.928$
(95\% CI $[0.896, 0.959]$), against $0.936$ predicted by
Equation~\ref{eq:auc}.}
\label{fig:discriminant}
\end{figure}

The primary development contrast, Set~3 versus Set~4, gave
$d' = 2.15$ (95\% CI $[1.83, 2.55]$) by Equation~\ref{eq:dprime}. The
two sets' interquartile ranges are disjoint
(Table~\ref{tab:by-set}). Mean scores differed by
$\Delta\mu = 0.244$ (95\% CI $[0.210, 0.276]$). The empirical area
under the ROC curve was $0.928$ (95\% CI $[0.896, 0.959]$), within
$0.008$ of the $0.936$ predicted by Equation~\ref{eq:auc}
(Figure~\ref{fig:discriminant}b). Confidence intervals throughout are
bootstrap percentile intervals over resampled source publications
($B = 5{,}000$; Section~\ref{sec:eval}).

The same contrast was recomputed on each of the five cluster scores
defined in Section~\ref{sec:metrics-agg} to localize where the
discriminant comes from (Table~\ref{tab:per-cluster-dprime}). Every
cluster separated the two sets on its own, from $d' = 0.72$ for pose
(C3) to $d' = 1.41$ for texture/sharpness (C2). The master exceeds the
strongest single cluster by $0.74$.

Supplementary Methods~\ref{sec:supp-s4} recomputes the Set~3 versus
Set~4 contrast under five alternative aggregation schemes
(Table~\ref{tab:agg-sensitivity}). Changing the weights moves the
discriminant little: equal cluster weights give $d' = 2.03$. Removing measurements costs more, with a reduced
three-metric score at $d' = 1.39$.

\begin{table}[h]
\centering
\caption{Master consistency score $S$ by comparison set.
$n$ denotes the number of pairs with a successfully computed master
score. $S \in [0, 1]$ with $1 = $ identical. Set~2 is summarized here
for completeness; its scores are read against perturbation magnitude in
Section~\ref{sec:results-weight-fit}
(Figure~\ref{fig:dose-response}), not as a level. Set~2's 32 unscored
pairs (Table~\ref{tab:set-composition}) are excluded from these summary
statistics.}
\label{tab:by-set}
\begin{tabular}{lrcccc}
\toprule
Set & $n$ & Mean & Median & SD & IQR \\
\midrule
1. Trivial positives          & 309 & 0.990 & 0.990 & 0.000 & [0.990, 0.990] \\
2. Sensitivity sweep          & 484 & 0.730 & 0.755 & 0.162 & [0.624, 0.865] \\
3. Real pre/post (within-pt.) & 134 & 0.504 & 0.517 & 0.126 & [0.418, 0.582] \\
4. Mismatched pairs           & 134 & 0.261 & 0.244 & 0.099 & [0.196, 0.328] \\
\bottomrule
\end{tabular}
\end{table}

\begin{table}[h]
\centering
\caption{Per-cluster discriminant performance on the Set~3 versus
Set~4 contrast ($n_{\text{Set~3}} = n_{\text{Set~4}} = 134$). Cluster
membership and the cluster weights $w_c$ of Equation~\ref{eq:master}
are defined in Sections~\ref{sec:metrics-agg}--\ref{sec:cluster-weights}. The
master row repeats the headline $d'$ for comparison. The weight
ordering and the $d'$ ordering differ; Section~\ref{sec:discussion-mechanism}
reads the two together. The Set~3 means sit near $0.500$ by
construction: the empirical-CDF backend
(Section~\ref{sec:metrics-jnd}) maps Set~3 to a uniform distribution.
C1 is the exception at $0.513$, because it holds all five Signal
Detection Theory sub-metrics.}
\label{tab:per-cluster-dprime}
\begin{tabular}{@{}lccccc@{}}
\toprule
Cluster & $w_c$ & Mean (Set~3) & Mean (Set~4) & $d'$ & 95\% CI \\
\midrule
C1 Photometric (6 sub-metrics)            & 0.339 & 0.513 & 0.294 & 1.39 & $[1.07, 1.76]$ \\
C2 Texture / sharpness (3 sub-metrics)    & 0.262 & 0.500 & 0.227 & 1.41 & $[1.09, 1.79]$ \\
C3 Pose (2 sub-metrics)                & 0.108 & 0.500 & 0.321 & 0.72 & $[0.38, 1.07]$ \\
C4 Illumination direction (1 sub-metric)  & 0.153 & 0.499 & 0.237 & 0.94 & $[0.69, 1.30]$ \\
C5 Pitch (1 sub-metric)                   & 0.138 & 0.500 & 0.220 & 1.03 & $[0.72, 1.43]$ \\
\midrule
Master $S$                                 & --- & 0.504 & 0.261 & 2.15 & $[1.83, 2.55]$ \\
\bottomrule
\end{tabular}
\end{table}

\subsection{Within-cohort grading of clinical pre/post pairs}\label{sec:results-within-cohort}

The master consistency score also grades pairs within Set~3 itself.
Scores there span $[0.18, 0.79]$, and Table~\ref{tab:by-set} gives the
distribution.

Splitting Set~3 into tertiles by master score and computing the mean
per-cluster score in each (Table~\ref{tab:set3-tertile-clusters})
identifies which clusters vary most across those tertiles. Every
cluster scores higher in the top tertile than in the bottom, from
$\Delta = +0.201$ for illumination direction (C4) to
$\Delta = +0.349$ for pitch (C5). Pose (C3) is the one cluster that does not rise
monotonically: its middle and top tertiles are level ($0.572$ and
$0.571$).

\begin{table}[h]
\centering
\caption{Mean cluster and master scores by Set~3 master-score tertile
($n = 45$, $44$ and $45$). Tertiles are formed from the master score,
and each cluster contributes to the split it is measured against.
$\Delta$ is the top-minus-bottom-tertile difference; larger values
identify the clusters whose variation tracks the master score within
the cohort. It is a raw mean difference and is not comparable with the
$d'$ of
Table~\ref{tab:per-cluster-dprime}, which divides by the pooled
standard deviation.}
\label{tab:set3-tertile-clusters}
\begin{tabular}{@{}lcccc@{}}
\toprule
Cluster                                       & Bottom & Middle & Top   & $\Delta$ \\
\midrule
C1 Photometric (6 sub-metrics)                & 0.387  & 0.530  & 0.621 & +0.234 \\
C2 Texture / sharpness (3 sub-metrics)        & 0.320  & 0.517  & 0.663 & +0.342 \\
C3 Pose (2 sub-metrics)                    & 0.359  & 0.572  & 0.571 & +0.212 \\
C4 Illumination direction (1 sub-metric)      & 0.404  & 0.488  & 0.604 & +0.201 \\
C5 Pitch (1 sub-metric)                       & 0.341  & 0.468  & 0.690 & +0.349 \\
\midrule
Master $S$                                    & 0.363  & 0.516  & 0.634 & +0.271 \\
\bottomrule
\end{tabular}
\end{table}

\subsection{Deployed analysis tool}\label{sec:results-deployed-tool}

\begin{figure}[h]
\centering
\includegraphics[width=0.85\linewidth]{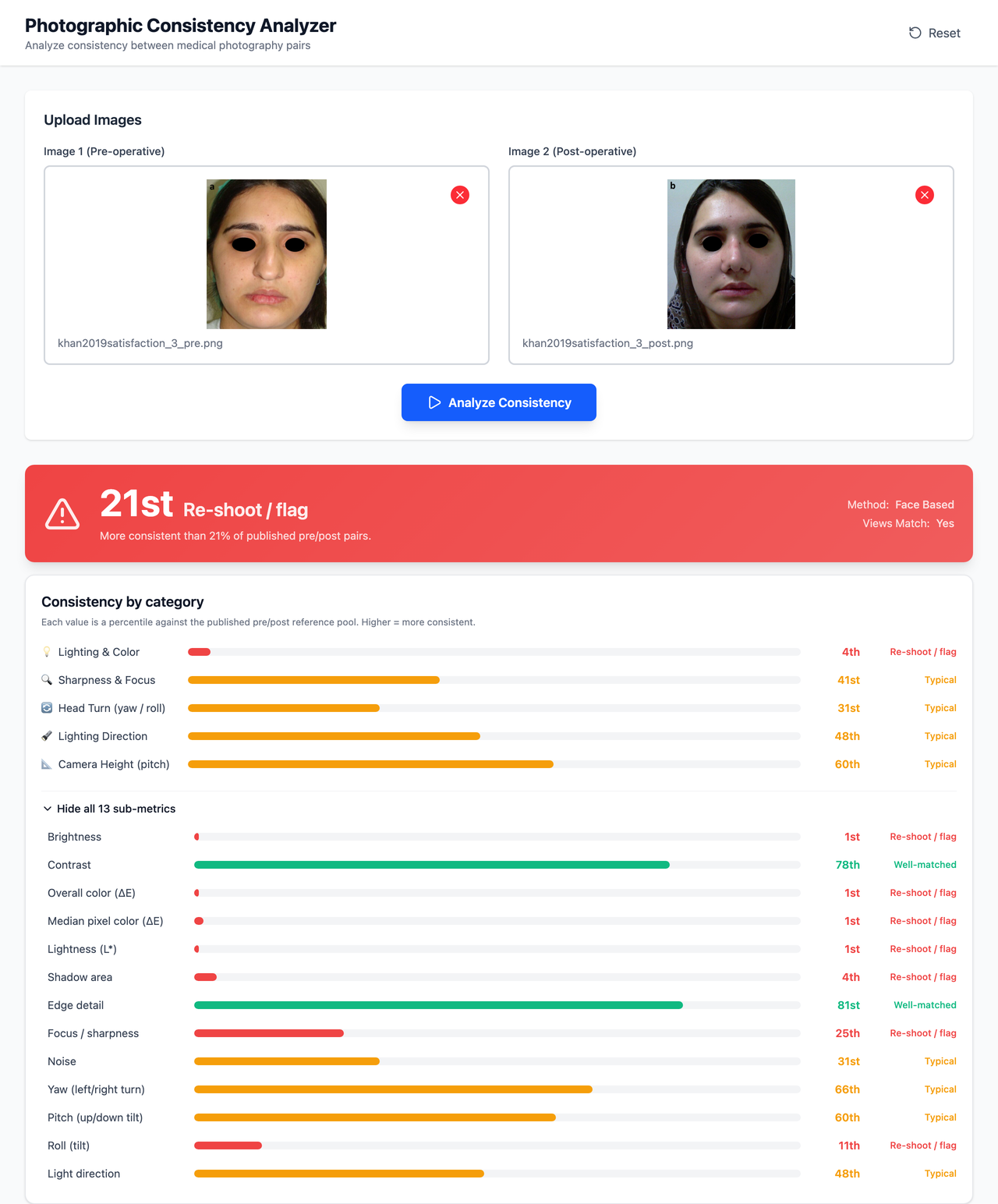}
\caption{Web interface for the photographic consistency tool,
single-pair result view. Scores are shown as percentiles of the Set~3
distribution; the pair illustrated sits at the 21st percentile and is
shown as an example of a pair the tool flags. The
interface labels the five clusters by their photographic meaning.
Against the tags of Table~\ref{tab:per-cluster-dprime}, Lighting \&
Color is C1, Sharpness \& Focus is C2, Head Turn is C3, Lighting
Direction is C4, and Camera Height is C5. Source pair reproduced
under CC~BY from \cite{khan2019satisfaction}.}
\label{fig:ui-screenshot}
\end{figure}

The pipeline of
Sections~\ref{sec:metrics}--\ref{sec:within-cohort-eval} is packaged as
a publicly accessible web tool
(Section~\ref{sec:implementation}; Figure~\ref{fig:ui-screenshot}). For
a single pair it reports the master score and the five cluster scores
as percentiles of the Set~3 distribution, together with the thirteen
sub-metrics, and it processes batches with live progress reporting. The interface
bands those percentiles into well-matched, typical, and re-shoot or
flag labels for workflow use. These bands are presentational
conveniences and not validated decision thresholds
(Section~\ref{sec:metrics}).

% =============================================================================
% Discussion
% =============================================================================

\section{Discussion}\label{sec:discussion}

\subsection{Principal findings}\label{sec:discussion-principal}
We developed a score for photographic consistency between paired
clinical images. Twelve of its 13 sub-metrics measure a capture
condition that published standards require to be held constant, and
noise is included on image-quality grounds
(Section~\ref{sec:metrics}). Correlation structure on within-patient pre/post pairs sorted the
sub-metrics into five clusters. The score is a weighted sum of the five cluster
scores.

The between-set contrast tests whether the score carries signal at
all. It separated within-patient pre/post pairs from pairs of unrelated
patients at $d' = 2.15$
(Section~\ref{sec:results-discriminant}). The within-cohort test holds identity fixed,
since every pair is one patient's own pre and post. Splitting Set~3 by
master score and reading the clusters back is a decomposition of that
score rather than an independent test of it: every cluster scored
higher in the top tertile than in the bottom, though pose did not rise
monotonically (Section~\ref{sec:results-within-cohort}). \chg{The perturbation sweep
alters one photograph along a single axis, holding the subject fixed;}
the score fell monotonically with magnitude
(Section~\ref{sec:results-weight-fit}). The separation is not an
artifact of leakage between publications or of the Gaussian model
behind $d'$
(Sections~\ref{sec:results-weight-fit}--\ref{sec:results-discriminant}).

\subsection{Comparison with prior literature}\label{sec:discussion-comparison}
\chg{None of the three prescriptive standards for clinical photography
supplies a metric by which a given pre/post pair can be checked
against the standard it nominally
meets}~\cite{DiBernardo1998,Prantl2017,ASPS_Guidelines}. That check is
still done by eye.

\chg{The tools to measure what those standards require already exist, each
built for another purpose.} CIEDE2000 quantifies perceived color
difference~\cite{Sharma2005}. Computer vision estimates
head pose from a single photograph~\cite{Hempel2022}, and
face-recognition programs publish tolerances for how far that pose may
drift between two images~\cite{NIST_FRVT_2024}. The sharpness and noise
metrics come from image-quality
research~\cite{Pertuz2013,DonohoJohnstone1994}; the direction of the
illuminant is recovered from the shading on a
face~\cite{Koenderink2003,Ramamoorthi2001,Zhou2015}. \chg{To our knowledge they have not been assembled into a single
image-derived score for the clinical pre/post pair.}

We assembled these existing measurements into thirteen sub-metrics for
the clinical pre/post pair. They arrive in incompatible units: a color
difference in $\Delta E_{00}$, a head rotation in degrees, a shadow
extent as a fraction of dark pixels, a noise difference as a log
ratio. \chg{We calibrated them onto one scale.} Each is
mapped to a score between 0 and 1, \chg{anchored so that each sub-metric's median across 134
within-patient pairs from 46 publications scores $0.5$}
(Section~\ref{sec:metrics-jnd}). A score of $0.4$ then represents the
same thing on every axis: less consistent than half of the published
pairs. We then grouped the calibrated scores by how closely they track
each other across those pairs (Section~\ref{sec:metrics-agg}).
Correlated sub-metrics would otherwise enter the sum as near-redundant
terms. The five cluster scores are combined into one master score.

\subsection{Mechanistic interpretation}\label{sec:discussion-mechanism}
The score breaks down along lines a photographer can act on. The five
clusters returned by the correlation structure of the 134
within-patient pairs (Section~\ref{sec:results-cluster-structure})
line up with five separate controls at capture: the level of the room
lighting and the white balance (C1), the camera and its settings (C2),
how the patient's head is turned (C3), where the light sits relative
to the face (C4), and the height of the camera (C5). The algorithm was
given the correlations between calibrated scores, with no labels and
no capture data. The alignment was not built in.

\chg{The split also indicates who holds each control.} The three head-pose
angles did not stay together: yaw and roll fell in one cluster, while
pitch stood alone. \chg{An instruction to the patient corrects yaw and roll.
Pitch depends on how high the photographer holds the camera.} Illumination direction also stood alone. A pair can fail
on either singleton while the rest looks normal. \chg{Each carries a full
cluster's weight; a flat average over the thirteen sub-metrics would
give it far less.}

The weights themselves matter little. Equal weights across the five
clusters work almost as well as the fitted ones
(Section~\ref{sec:results-weight-fit}).

\subsection{Score interpretation}\label{sec:discussion-interpretation}
\chg{Three reference points fix what a score means.}
Set~1 pairs each photograph with itself and sits at $S = 0.990$. Set~4
pairs photographs drawn from different publications and sits at
$0.261$, which is what the pipeline returns when two captures match
only by chance. Real pre/post pairs (Set~3) sit between the two,
averaging $0.504$ (Table~\ref{tab:by-set}).

A real pre/post pair cannot reach the identity ceiling. The surgery
itself moves some of the sub-metrics: swelling shifts skin color, and
a reshaped nose casts a different shadow. The best pair in the pool
scored $0.793$ (Section~\ref{sec:results-discriminant}). \chg{Capture differences move the same sub-metrics, and the score does
not say how a given gap divides between the two.} \chg{A pair can also be consistent and
still be two poor photographs, which this score does not address.}

\subsection{Clinical implications}\label{sec:discussion-clinical}
The five clusters suggest a capture checklist. \chg{No capture practice was
observed in this study; each item below follows from what its cluster
tracks.} C1 covers the photographic environment: the room lighting, the
backdrop, and the white balance. C2 covers the optics: the camera, its
distance from the subject, and its settings. C3 covers what a verbal
cue can address, since yaw and roll follow from how the patient holds
the head. C4 and C5 belong to the photographer, who sets where the
light falls and how high the camera sits.

During the post-operative visit, the score can be computed against the
archived pre-operative photograph while a re-shoot is still possible.
Before submission, a surgeon can audit a pair the same way. For an
existing archive, the score ranks pairs for inclusion in a case series
or an outcome study.

A reader who did not take the photographs faces the same question: how
much of the visible change belongs to the procedure. The score makes
that question explicit. A pair near the middle of the published
distribution carries the capture variability of most of the
literature. \chg{A lower-scoring pair differs more, and the
photographs alone cannot say how much of that difference is the
procedure.} The pipeline reports percentiles against that pool
(Section~\ref{sec:results-deployed-tool}).

\subsection{Strengths}\label{sec:discussion-strengths}
\chg{A disagreement with a master score can be traced to a cluster, from
there to a sub-metric, and from there to the calibrated tolerances
behind it} (Supplementary Methods~\ref{sec:supp-s3}). Those tolerances are fit on real
within-patient pairs and cross-checked against published photographic
and color-science just-noticeable differences. The five clusters carry
the discriminant on their own, \chg{so equal weights are a workable
default where refitting is not possible} (Section~\ref{sec:discussion-mechanism}). \chg{The pipeline
runs as a web tool that keeps nothing: uploaded pairs are deleted at
the end of the request and no image bytes are written to disk or to
logs} (Section~\ref{sec:implementation}).

\subsection{Limitations and future work}\label{sec:discussion-limitations}
Three limitations qualify these results, and each points at the work
that would settle it.

The ground truth is a proxy. Set~3 versus Set~4 tests whether the
score separates real pairs from unrelated ones, and \chg{no reader study
asked a clinician whether a pair is acceptable, so this study reports
no operating cutoffs and recommends scoring within a cohort.} Surgeon
raters, blinded to the score, grading held-out pre/post pairs on
lighting, focus and pose would replace the proxy and show where the
calibration departs from expert judgment. \chg{The contrast also conflates
capture style with patient identity; the 35 publications contributing
more than one Set~3 pair (Table~\ref{tab:corpus-pubs}) yield
224 same-publication, different-patient combinations, enough for a size-matched control set
that would separate the two.}

\chg{The cluster weights were fitted by maximizing $d'$ on the same Set~3
versus Set~4 contrast the headline reports;} cross-validated and
publication-level held-out values are given in
Section~\ref{sec:results-weight-fit} and Supplementary
Methods~\ref{sec:supp-s4}. A future version could drop the fit,
and the circularity with it (Section~\ref{sec:discussion-mechanism}).

All photographs are single frontal views published between
2010 and 2025 in US-based journals or with an English translation, \chg{and five published capture requirements have no
sub-metric at all, among them subject-to-camera distance and facial
expression} (Supplementary Methods~\ref{sec:supp-provenance}). Lateral, oblique and intraoral views need a pose-aware
alignment stage and a calibration of their own. \chg{Every contributing publication is facial
surgery, so the pose sub-metrics assume a head; reaching breast or
body contouring would take different measurements.} Applying the
pipeline outside the US would test geographic transfer.

\section{Conclusion}\label{sec:conclusion}
We built a score for how consistently a pre/post pair was captured,
from thirteen sub-metrics grouped into five clusters discovered in the
correlation structure of published pairs. On the development contrast
it separated real within-patient pairs from mismatched pairs at
$d' = 2.15$. A reader study is the next step, and
no operating threshold should be set before it.

% =============================================================================
% Disclosure of generative AI use (SPIE/ICMJE: AI cannot be an author; tool use
% in writing must be disclosed. A-type "language editing" statement.)
% =============================================================================
\section*{Disclosures}
The authors declare no conflicts of interest.

\chg{\section*{Code, Data, and Materials Availability}\label{sec:availability}
The analysis code, fitted cluster weights, and the declared dependency
requirements used to produce all results reported here are available
from the corresponding author upon reasonable request. The deployed web
application is publicly hosted at
\url{https://rick95125-photographic-consistency.hf.space}. All
photographs analysed were extracted from published articles; the
contributing publications are listed in Supplementary
Methods~\ref{sec:supp-s1}.}

\chg{\section*{Ethics Statement}
This study analysed photographs already published in the peer-reviewed
literature and did not involve interaction with human subjects or
access to identifiable private information, and so does not constitute
human-subjects research under 45~CFR~46.102(e)
(Section~\ref{sec:study-design}). Figure~\ref{fig:s1-perturbation-panel}
reproduces an image published under a Creative Commons Attribution
licence, with the licence and the modification stated in the caption.}

\chg{\section*{Acknowledgments}
No funding supported this work. During the preparation of this
manuscript the authors used large language models (ChatGPT, OpenAI;
Claude, Anthropic) to improve readability and language. After using
these tools the authors reviewed and edited the content and take full
responsibility for the content of the publication.}

% =============================================================================
% References
% =============================================================================
\bibliographystyle{spiejour}
\bibliography{references,references_corpus}

% =============================================================================
% Supplementary Materials
% =============================================================================
% =============================================================================
%  Supplementary Methods, Path A pilot manuscript
% =============================================================================

\clearpage
\appendix
\section*{Supplementary Methods}
\renewcommand{\thetable}{S\arabic{table}}
\setcounter{table}{0}
\renewcommand{\thefigure}{S\arabic{figure}}
\setcounter{figure}{0}
\renewcommand{\thesubsection}{S\arabic{subsection}}
\setcounter{subsection}{0}
\renewcommand{\theequation}{S\arabic{equation}}
\setcounter{equation}{0}

\subsection{Source photographs: provenance, inclusion attrition, and Set~2 perturbation axes}
\label{sec:supp-s1}

\paragraph{PubMed search string.}
The photograph pool was retrieved from PubMed on 1 August 2025
(publications dated 2010--2025) with the following Boolean query, applied to the
Title/Abstract fields:
\begin{verbatim}
("facial plastic surgery"[Title/Abstract] OR "aesthetic surgery"[Title/Abstract]
 OR rhinoplasty[Title/Abstract] OR facelift[Title/Abstract]
 OR blepharoplasty[Title/Abstract])
AND
("before and after"[Title/Abstract] OR "clinical photography"[Title/Abstract]
 OR "standardized photography"[Title/Abstract]
 OR "photographic documentation"[Title/Abstract])
AND
(methods[Title/Abstract] OR methodology[Title/Abstract]
 OR "photo bias"[Title/Abstract] OR "photographic bias"[Title/Abstract]
 OR lighting[Title/Abstract] OR expression[Title/Abstract]
 OR framing[Title/Abstract] OR makeup[Title/Abstract]
 OR positioning[Title/Abstract])
\end{verbatim}

\paragraph{Source publications.}
Search yield and per-stage inclusion attrition are reported in
Section~\ref{sec:results-corpus} (Figure~\ref{fig:prisma}). Among the
52 publications that contributed at least one photograph clearing all
inclusion gates, the per-study yield ranged from 1 to 18 photographs
(median $4$, mean $5.9$); five publications contributed a single
photograph, 23 contributed 2--5, 18 contributed 6--10, and six
contributed more than 10. The full per-publication characteristics, journal, publication year, procedure type, study design, and the number of
photographs and within-patient pre/post pairs contributed, are listed in
Table~\ref{tab:corpus-pubs}.

% Auto-generated: Photo1_Literatures.xlsx (journals/years/DOIs) + corpus counts +
% PubMed-verified procedures (corpus_procedure_verified.csv) + PubMed study designs.
% Requires \usepackage{longtable}, \usepackage{booktabs}.
\footnotesize
\begin{longtable}{@{}l p{2.8cm} l l c c@{}}
\caption{Characteristics of the included publications contributing facial pre/post photographs to the photograph pool. ``Photographs'' is the number of eligible photographs contributed and ``Pairs'' the number of real within-patient pre/post image pairs (Set~3). Procedure type was verified against each article; study design is classified from PubMed publication types and abstract design language.}\label{tab:corpus-pubs}\\
\toprule
First author (year) & Journal & Procedure & Design & Photos & Pairs \\
\midrule
\endfirsthead
\multicolumn{6}{c}{\tablename\ \thetable{} -- continued from previous page}\\
\toprule
First author (year) & Journal & Procedure & Design & Photos & Pairs \\
\midrule
\endhead
\midrule \multicolumn{6}{r}{\textit{continued on next page}}\\
\endfoot
\endlastfoot
Agarwal 2012~\cite{agarwal2012alar} & J Craniofac Surg & Rhinoplasty/nasal & Case series & 5 & 2 \\
Akpolat 2022~\cite{akpolat2022effect} & J Cosmet Dermatol & Rhinoplasty/nasal & Prospective & 2 & 1 \\
Amali 2014~\cite{amali2014assessment} & Aesthet Surg J & Rhinoplasty/nasal & RCT & 4 & 2 \\
Apaydin 2011~\cite{apaydin2011nasal} & Facial Plast Surg & Rhinoplasty/nasal & Case series & 6 & 3 \\
Barone 2024~\cite{barone2024reconstruction} & Aesthetic Plast Surg & Rhinoplasty/nasal & RCT & 2 & 1 \\
Dorfman 2020~\cite{dorfman2020making} & Aesthet Surg J & Rhinoplasty/nasal & Retrospective & 8 & 4 \\
Elkashty 2023~\cite{elkashty2023outcomes} & Ann Maxillofac Surg & Rhinoplasty/nasal & Case series & 7 & 3 \\
Eren 2014~\cite{eren2014autospreading} & Aesthetic Plast Surg & Rhinoplasty/nasal & Case series & 6 & 2 \\
Gassling 2015~\cite{gassling2015secondary} & J Craniomaxillofac Surg & Rhinoplasty/nasal & Case series & 4 & 2 \\
Hafezi 2020~\cite{hafezi2020practical} & Plast Reconstr Surg & Rhinoplasty/nasal & Case series & 2 & 1 \\
Khan 2019~\cite{khan2019satisfaction} & Cureus & Rhinoplasty/nasal & Case series & 6 & 3 \\
Kim 2015~\cite{kim2015use} & Plast Reconstr Surg & Rhinoplasty/nasal & Retrospective & 7 & 3 \\
Kosins 2015~\cite{kosins2015plunging} & Aesthet Surg J & Rhinoplasty/nasal & Prospective & 4 & 2 \\
Lee 2015~\cite{lee2015columella} & Arch Plast Surg & Rhinoplasty/nasal & Case report & 1 & 0 \\
Lee 2018~\cite{lee2018nasal} & Aesthetic Plast Surg & Rhinoplasty/nasal & Case series & 3 & 1 \\
Lee 2022~\cite{lee2022crushed} & Aesthetic Plast Surg & Rhinoplasty/nasal & Retrospective & 18 & 9 \\
Marianetti 2025~\cite{marianetti2025alar} & Aesthetic Plast Surg & Rhinoplasty/nasal & Case series & 3 & 1 \\
Murrell 2014~\cite{murrell2014correlation} & Aesthet Surg J & Rhinoplasty/nasal & Prospective & 3 & 1 \\
Nguyen 2022~\cite{nguyen2022bony} & J Craniofac Surg & Rhinoplasty/nasal & Case series & 10 & 5 \\
Ozturan 2015~\cite{ozturan2015wide} & Aesthetic Plast Surg & Rhinoplasty/nasal & Case series & 1 & 0 \\
Ozturk 2020~\cite{ozturk2020new} & Aesthetic Plast Surg & Rhinoplasty/nasal & Case series & 18 & 9 \\
Parsa 2021~\cite{parsa2021role} & Aesthetic Plast Surg & Rhinoplasty/nasal & Retrospective & 4 & 2 \\
Plath 2023~\cite{plath2023does} & Eur Arch Otorhinolaryngol & Rhinoplasty/nasal & Prospective & 7 & 3 \\
Radulesco 2022~\cite{radulesco2022liquid} & Aesthetic Plast Surg & Rhinoplasty/nasal & Case series & 1 & 0 \\
Radulesco 2024~\cite{radulesco2024prospective} & Aesthetic Plast Surg & Rhinoplasty/nasal & Prospective & 6 & 3 \\
Roxbury 2012~\cite{roxbury2012impact} & Laryngoscope & Rhinoplasty/nasal & RCT & 2 & 1 \\
Sazgar 2021~\cite{sazgar2021different} & Braz J Otorhinolaryngol & Rhinoplasty/nasal & Prospective & 7 & 3 \\
Shuaib 2015~\cite{shuaib2015can} & Plast Reconstr Surg & Rhinoplasty/nasal & Retrospective & 12 & 6 \\
Suh 2021~\cite{suh2021change} & Arch Craniofac Surg & Rhinoplasty/nasal & Case series & 4 & 2 \\
Tanaka 2014~\cite{tanaka2014oriental} & Plast Reconstr Surg Glob Open & Rhinoplasty/nasal & Case series & 4 & 2 \\
Tiong 2014~\cite{tiong2014augmentation} & Plast Surg Int & Rhinoplasty/nasal & Retrospective & 1 & 0 \\
Wang 2023~\cite{wang2023clinical} & Zhonghua Er Bi Yan Hou Tou Jing Wai Ke Za Zhi & Rhinoplasty/nasal & Retrospective & 6 & 3 \\
Wei 2018~\cite{wei2018use} & Ann Plast Surg & Rhinoplasty/nasal & Retrospective & 3 & 1 \\
Yen 2021~\cite{yen2021comparative} & Aesthetic Plast Surg & Rhinoplasty/nasal & Retrospective & 4 & 2 \\
Zheng 2023~\cite{zheng2023improvement} & Aesthetic Plast Surg & Rhinoplasty/nasal & Case series & 13 & 6 \\
Bitik 2020~\cite{bitik2020sub} & Aesthet Surg J & Facelift/midface & Retrospective & 14 & 6 \\
Bitik 2023~\cite{bitik2023intraorbital} & Aesthet Surg J & Facelift/midface & Retrospective & 18 & 8 \\
Hoenig 2011~\cite{hoenig2011vertical} & Aesthetic Plast Surg & Facelift/midface & Case series & 9 & 2 \\
Sirinoglu 2025~\cite{csirinoglu2025long} & J Craniofac Surg & Facelift/midface & Retrospective & 10 & 5 \\
Weng 2019~\cite{weng2019achieving} & Aesthet Surg J & Facelift/midface & Retrospective & 8 & 4 \\
Boonipat 2022~\cite{boonipat2022impact} & Plast Reconstr Surg & Periorbital & Case series & 6 & 3 \\
Kesty 2025~\cite{kesty2025novel} & J Cosmet Dermatol & Periorbital & Case report & 2 & 1 \\
Kim 2013~\cite{kim2013effects} & Aesthetic Plast Surg & Periorbital & Retrospective & 2 & 0 \\
Park 2024~\cite{park2024choosing} & Aesthetic Plast Surg & Periorbital & Retrospective & 10 & 2 \\
Patrocinio 2011~\cite{patrocinio2011complications} & Braz J Otorhinolaryngol & Periorbital & Retrospective & 4 & 2 \\
Soares 2018~\cite{soares2018effect} & Arq Bras Oftalmol & Periorbital & Prospective & 2 & 1 \\
Feinendegen 2016~\cite{feinendegen2016brow} & J Craniomaxillofac Surg & Brow & Case series & 4 & 2 \\
Isik 2012~\cite{isik2012contour} & Aesthetic Plast Surg & Forehead & Case series & 3 & 1 \\
Lots 2023~\cite{lots2023effect} & J Cosmet Dermatol & Injectable & Case series & 1 & 0 \\
Sterodimas 2025~\cite{sterodimas2025prospective} & Plast Reconstr Surg Glob Open & Injectable & Prospective & 8 & 3 \\
Taub 2010~\cite{taub2010effect} & Dermatol Surg & Injectable & Case series & 10 & 3 \\
Barrera 2018~\cite{barrera2018efficacy} & Plast Reconstr Surg Glob Open & Other & Retrospective & 4 & 2 \\
\midrule
\textbf{Total} & & & & \textbf{309} & \textbf{134} \\
\bottomrule
\end{longtable}
\normalsize

% [Moved to outcome paper 2026-06-07] The representative-pair score breakdown
% paragraph and \input{example_subscores_table} referenced fig:examples /
% sec:results-examples, which migrated to the outcome paper with the
% surgeon-facing results. Removed here so the pipeline paper's refs resolve.

\paragraph{Set~2 perturbation axes.}
Set~2 (sensitivity sweep, Section~\ref{sec:comparison-sets}) consists of
single-axis perturbations applied independently to 12 source photographs
drawn from the eligible pool. Table~\ref{tab:s1-set2-axes} lists the five
perturbation axes, their units, and the magnitude values applied; each of
the 12 source photographs was perturbed at every listed magnitude on every
axis, producing $12 \times 43 = 516$ source-vs-perturbed pairs.

\begin{table}[H]
  \centering
  \footnotesize
  \caption{Set~2 perturbation axes. Each of 12 source photographs was
  perturbed independently along every axis at every listed magnitude,
  producing $12 \times 43 = 516$ source-vs-perturbed pairs.}
  \label{tab:s1-set2-axes}
  \begin{tabular}{@{}llcl@{}}
    \toprule
    \textbf{Axis} & \textbf{Unit} & \textbf{\# Levels} & \textbf{Magnitude values} \\
    \midrule
    Gaussian blur     & $\sigma$ (px)     & 5  & $0.5,\ 1,\ 2,\ 4,\ 8$ \\
    Brightness        & \% offset         & 8  & $\pm 2,\ \pm 10,\ \pm 25,\ \pm 50$ \\
    White balance     & $\Delta K$        & 10 & $\pm 50,\ \pm 200,\ \pm 500,\ \pm 1000,\ \pm 2000$ \\
    Rotation          & degrees           & 12 & $\pm 0.5,\ \pm 2,\ \pm 5,\ \pm 10,\ \pm 20,\ \pm 30$ \\
    Scale             & factor            & 8  & $0.90,\ 0.95,\ 0.97,\ 0.99,\ 1.01,\ 1.03,\ 1.05,\ 1.10$ \\
    \midrule
    \textbf{Total}    &                   & \textbf{43} & \\
    \bottomrule
  \end{tabular}
\end{table}

\begin{figure}[H]
\centering
\includegraphics[width=0.9\linewidth]{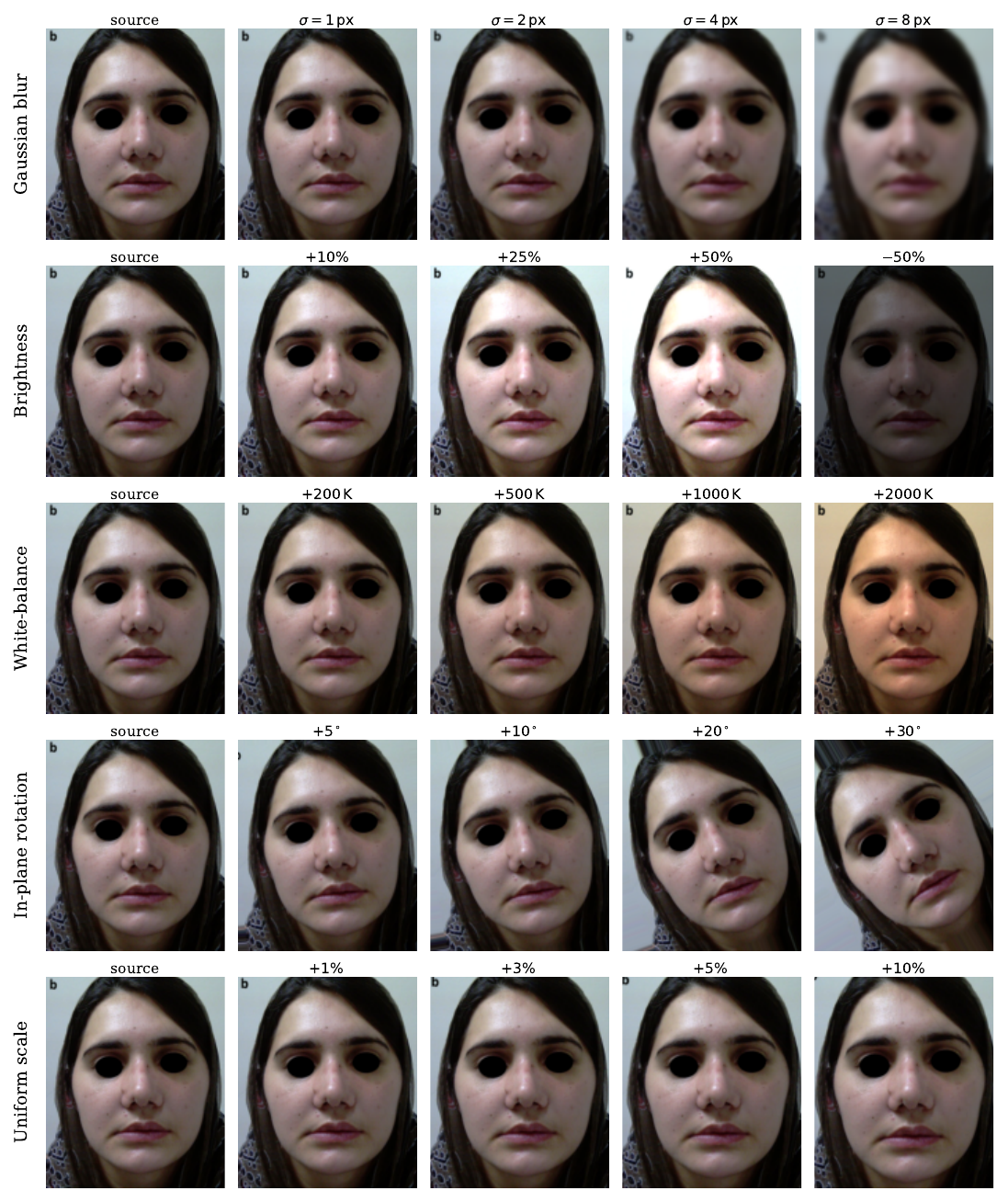}
\caption{Set~2 perturbation panel for a representative source photograph
(\emph{khan2019satisfaction\_3\_post}), reproduced and modified under
CC~BY from \cite{khan2019satisfaction}. Rows show the five
single-axis perturbations applied independently in Set~2; columns show
the unperturbed source alongside four increasing-magnitude perturbations
sampled from the magnitude grid of Table~\ref{tab:s1-set2-axes}. The
brightness row's final column shows $-50\%$ to demonstrate the symmetric
response to bilateral perturbations. The smaller magnitudes shown
(e.g., $\sigma{=}1$\,px blur, $+200$\,K white-balance shift, $+1\%$
scale) are near-imperceptible at this print size. \chg{Set~2 includes
magnitudes below and above the visible threshold, and is not used to
fit the cluster weights
(Section~\ref{sec:results-weight-fit}).}
The rotation operator preserves canvas dimensions
($641 \times 753$\,px), so at $+30^{\circ}$ the original-image corners
rotate out of frame and the new corners are filled with content from
outside the original bounds.}
\label{fig:s1-perturbation-panel}
\end{figure}

\subsection{Photo-level inclusion gates: thresholds and detection libraries}
\label{sec:supp-s2}

\chg{Table~\ref{tab:s2-gates} gives the measurement, exclusion threshold,
and implementing library for each of the six gates described in
Section~\ref{sec:inclusion}. A photograph is attributed to the first
gate it fails and is not evaluated against the rest. Static gates read
the BGR image returned by OpenCV \texttt{cv2.imread}, and HSV
statistics use the OpenCV scale ($S, V \in [0, 255]$). Per-gate
attrition is reported in Section~\ref{sec:results-corpus}.}

\begin{table}[H]
  \centering
  \footnotesize
  \caption{Photo-level inclusion gates. Order is fixed; a photograph is
  attributed to the first gate it fails.}
  \label{tab:s2-gates}
  \begin{tabular}{@{}p{20mm}p{55mm}p{40mm}p{30mm}@{}}
    \toprule
    \textbf{Gate} & \textbf{Measurement} & \textbf{Exclusion threshold} & \textbf{Library} \\
    \midrule
    Too small &
      $\min(w, h)$ in pixels &
      $< 300$ &
      OpenCV \\
    \addlinespace
    Grayscale / low color &
      Mean HSV saturation $\bar{S}$ &
      $\bar{S} \le 12.0$ &
      OpenCV \\
    \addlinespace
    Exposure problem &
      Fraction of near-black pixels ($V < 10$) \emph{or} fraction of saturated-highlight pixels ($V > 245$) &
      near-black fraction $\ge 0.15$ \emph{or} highlight fraction $\ge 0.10$ &
      OpenCV \\
    \addlinespace
    Text overlay &
      Canny edge-pixel fraction (fixed thresholds $100, 200$) &
      $\ge 0.060$ &
      OpenCV \\
    \addlinespace
    Face undetected &
      MediaPipe Face Detector landmark extraction succeeds on the highest-confidence detection &
      no face passes
      $\text{min\_detection\_confidence} = 0.5$ &
      MediaPipe Tasks~\cite{Lugaresi2019} \\
    \addlinespace
    Pose unrecoverable &
      6DRepNet head-pose estimation succeeds &
      estimator returns failure &
      6DRepNet~\cite{Hempel2022} \\
    \bottomrule
  \end{tabular}
\end{table}

\chg{The four static gates run on every photograph. The two model-based
gates run only on photographs that clear all four, because MediaPipe
and 6DRepNet inference costs far more than an image statistic. The
four static thresholds were fixed in advance and were not tuned
against any result reported here.} \chg{The gate implementation is covered by the code availability
statement in the end matter.}

\subsection{Sub-metric calibration: transforms, backends, and fitted JNDs}
\label{sec:supp-s3}

Each of the thirteen calibrated sub-metrics maps a raw between-photograph
difference to a $[0, 1]$ consistency score through a two-step pipeline. The
transform and backend are fixed at calibration time. \chg{Every result in
this paper uses one frozen calibration, fit on Set~3 ($n = 134$ real
pre/post pairs) and never refit thereafter.}

\paragraph{Step 1, variance-stabilizing transform.}
The absolute raw difference $x \ge 0$ is passed through a metric-specific monotone
transform $T(\cdot)$ chosen by inspection of the Set~3 distribution and the
underlying perceptual framework:
\begin{itemize}
  \item \chg{\textbf{Square-root} ($T(x) = \sqrt{x}$) for lightness
        differences, on the CIELAB $L^{*}$ axis and on HSV
        $V$~\cite{CIE15_2004,Mahy1994}.}
  \item \chg{\textbf{$\ln(1 + x)$} for differences with a heavy upper tail:
        the two $\Delta E_{00}$ color differences and the contrast
        difference.}
  \item \textbf{Identity} ($T(x) = x$) for metrics that already lie on a bounded
        or near-symmetric scale (pose angle differences, log-ratio quality
        metrics, normalized illumination-direction cosine distance).
\end{itemize}

\paragraph{Step 2, backend mapping.}
The transformed difference $T(x)$ is mapped to a score by one of two backends.

\textit{Signal Detection Theory (SDT) backend.} For metrics whose transformed
distribution is approximately Gaussian under Set~3 real pairs, we use the
classical SDT formulation~\cite{GreenSwets1966}:
\begin{equation}
  d'_{m} \;=\; \frac{T(x)}{\mathrm{JND}_{t}}, \qquad
  \mathrm{score}(x) \;=\; 2\,\Phi^{\mathrm{c}}\!\left( \tfrac{d'_{m}}{2} \right),
  \label{eq:s3-sdt}
\end{equation}
where $\Phi^{\mathrm{c}}(z) = 1 - \Phi(z)$ is the complementary Gaussian
cumulative distribution function and $\mathrm{JND}_{t}$ is the metric's fitted
Just-Noticeable-Difference in transformed units. The per-pair, per-metric
quantity $d'_{m}$ is written with a subscript to distinguish it from the
between-set discriminant $d'$ of main-text Equation~\ref{eq:dprime}.
$\mathrm{JND}_{t}$ is chosen empirically so that the median Set~3 real pre/post
pair lands at $d'_{m} = 2\,\Phi^{-1}(0.75) \approx 1.349$, the sensitivity
index at which a two-alternative forced-choice observer is $75\%$
correct~\cite{GreenSwets1966,MacmillanCreelman2005}; this anchors the
median score at $0.5$ by construction. Equivalently,
$\mathrm{JND}_{t} = T(x_{\mathrm{med}}) / 2\,\Phi^{-1}(0.75)$, where
$x_{\mathrm{med}}$ is that sub-metric's median raw difference on Set~3.

\textit{Empirical CDF (ECDF) backend.} For metrics whose transformed distribution
remains non-Gaussian (heavy-tailed, bimodal, or supported on a bounded interval),
we replace the SDT mapping with a direct empirical-CDF lookup against the Set~3
real pre/post distribution:
\begin{equation}
  \mathrm{score}(x) \;=\; 1 - \widehat{F}(x),
  \label{eq:s3-ecdf}
\end{equation}
where $\widehat{F}$ is the linearly-interpolated empirical CDF built from the
sorted Set~3 values with Hazen plotting positions
$\widehat{F}(x_{(k)}) = (k - \tfrac{1}{2})/n$. The ECDF backend requires no
$\mathrm{JND}$ parameter and is monotone-invariant, so no transform is needed
($T(x) = x$).

\paragraph{Per-metric assignments.}
Table~\ref{tab:s3-calibration} reports, for each of the thirteen sub-metrics:
its cluster assignment, raw measurement, transform, backend, fitted
$\mathrm{JND}_{t}$ (transformed units, SDT only), and the perceptual or
biometric references that motivate the measurement and serve as cross-checks on
the fitted threshold. \chg{The cluster weights are fit separately, on the same Set~3, and are
reported in Section~\ref{sec:results-weight-fit}.}

\begin{table}[H]
  \centering
  \scriptsize
  \caption{Per-metric calibration. \emph{Transform} $T(\cdot)$ is applied to the
  raw absolute difference; \emph{Backend} selects between Equation~\ref{eq:s3-sdt}
  (SDT) and Equation~\ref{eq:s3-ecdf} (ECDF). \emph{$\mathrm{JND}_{t}$} is the fitted
  JND in transformed units; \chg{a dash indicates the ECDF backend, which is
  parameter-free.} All fits use Set~3 ($n = 134$).}
  \label{tab:s3-calibration}
  \begin{tabular}{@{}p{16mm}p{30mm}p{32mm}p{10mm}p{8mm}p{12mm}p{20mm}@{}}
    \toprule
    \textbf{Cluster} & \textbf{Sub-metric} & \textbf{Raw measurement} &
      \textbf{$T(\cdot)$} & \textbf{Backend} & \textbf{$\mathrm{JND}_{t}$} &
      \textbf{Refs} \\
    \midrule
    C1 Photometric & brightness\_diff & $|\,\bar{V}_{\mathrm{pre}} - \bar{V}_{\mathrm{post}}\,|$ on $V$ (HSV) & $\sqrt{x}$ & SDT & $2.209$ & \cite{CIE15_2004,Mahy1994,Weber1834} \\
    \addlinespace
    C1 Photometric & contrast\_diff & $|\,\sigma_{V,\mathrm{pre}} - \sigma_{V,\mathrm{post}}\,|$ & $\ln(1{+}x)$ & SDT & $1.276$ & \cite{Weber1834} \\
    \addlinespace
    C1 Photometric & shadow\_area\_diff & $|\,A_{\mathrm{shadow,pre}} - A_{\mathrm{shadow,post}}\,|$ (frac.\ of pixels with $V<$ adaptive threshold) & $x$ & ECDF & --- & \cite{CIE15_2004} \\
    \addlinespace
    C1 Photometric & global\_delta\_e & Mean CIEDE2000 $\Delta E_{00}$ over aligned skin region & $\ln(1{+}x)$ & SDT & $1.292$ & \cite{Sharma2005,LuoRigg1986} \\
    \addlinespace
    C1 Photometric & median\_pixel\_delta\_e & Median per-pixel CIEDE2000 $\Delta E_{00}$ over aligned skin region & $\ln(1{+}x)$ & SDT & $1.422$ & \cite{Sharma2005,Melgosa1997} \\
    \addlinespace
    C1 Photometric & lab\_l\_mean\_diff & $|\,\bar{L}^{*}_{\mathrm{pre}} - \bar{L}^{*}_{\mathrm{post}}\,|$ in CIELAB & $\sqrt{x}$ & SDT & $2.259$ & \cite{CIE15_2004,Mahy1994} \\
    \addlinespace
    C2 Texture & gradient\_diff & $|\,\overline{|\nabla I|}_{\mathrm{pre}} - \overline{|\nabla I|}_{\mathrm{post}}\,|$ (Sobel magnitude) & $x$ & ECDF & --- & \cite{Pertuz2013} \\
    \addlinespace
    C2 Texture & focus\_lapv\_log\_ratio & $|\ln(\mathrm{LAPV}_{\mathrm{pre}}/\mathrm{LAPV}_{\mathrm{post}})|$ (variance of Laplacian) & $x$ & ECDF & --- & \cite{Pertuz2013} \\
    \addlinespace
    C2 Texture & noise\_estimate\_log\_ratio & $|\ln(\hat{\sigma}_{\mathrm{pre}}/\hat{\sigma}_{\mathrm{post}})|$ (MAD noise estimator) & $x$ & ECDF & --- & \cite{DonohoJohnstone1994} \\
    \addlinespace
    C3 Pose & yaw\_diff & $|\,\mathrm{yaw}_{\mathrm{pre}} - \mathrm{yaw}_{\mathrm{post}}\,|$ (deg) from 6DRepNet & $x$ & ECDF & --- & \cite{Hempel2022,NIST_FRVT_2024,Wilson2000} \\
    \addlinespace
    C5 Pitch & pitch\_diff & $|\,\mathrm{pitch}_{\mathrm{pre}} - \mathrm{pitch}_{\mathrm{post}}\,|$ (deg) from 6DRepNet & $x$ & ECDF & --- & \cite{Hempel2022,NIST_FRVT_2024,Wilson2000} \\
    \addlinespace
    C3 Pose & roll\_diff & $|\,\mathrm{roll}_{\mathrm{pre}} - \mathrm{roll}_{\mathrm{post}}\,|$ (deg) from 6DRepNet & $x$ & ECDF & --- & \cite{Hempel2022,NIST_FRVT_2024,Wilson2000} \\
    \addlinespace
    C4 Illum. dir. & light\_direction\_diff & Angular distance between estimated dominant illuminant directions (rad) & $x$ & ECDF & --- & \cite{Koenderink2003,Zhou2015,Ramamoorthi2001} \\
    \bottomrule
  \end{tabular}
\end{table}

\paragraph{Cross-checks against literature thresholds.}
The literature-derived thresholds referenced in the main text appear in this work
as motivation for the transform choice and as sanity checks on the fitted JND,
not as the JND values themselves. For the five SDT metrics, the fitted
$\mathrm{JND}_{t}$ corresponds (after inverting the transform) to raw differences
in the range conventionally judged perceptible by an attentive observer
($\mathrm{JND}_{t}=2.209$ on $\sqrt{\cdot}$ for brightness corresponds to a raw
$V$-difference of $\approx 4.9$ on $[0, 255]$, comfortably above the
Weber-fraction floor~\cite{Weber1834}; $\mathrm{JND}_{t}=1.292$ on $\ln(1{+}\cdot)$
for global $\Delta E_{00}$ corresponds to $\approx 2.64$ $\Delta E$ units, in the
range identified by Sharma~\cite{Sharma2005} and Luo \& Rigg~\cite{LuoRigg1986}
as just-perceptible). For the eight ECDF metrics, NIST FRVT pose
tolerances~\cite{NIST_FRVT_2024} and Wilson \emph{et al.}~\cite{Wilson2000}
provide qualitative anchors for the rotational angles, and the
Koenderink~\cite{Koenderink2003} / Zhou~\cite{Zhou2015} /
Ramamoorthi~\cite{Ramamoorthi2001} bodies of work motivate the use of an
illumination-direction discrepancy rather than a color-only summary.

\subsection{Cluster construction and weight fitting}
\label{sec:supp-s4}

This section gives the full specification of the two-stage aggregation
referenced in main-text Sections~\ref{sec:metrics-agg}
and~\ref{sec:cluster-weights}: first, how the thirteen calibrated
sub-metrics are partitioned into five clusters, and second, how the
five cluster weights are fit. The final model uses Set~3-derived
correlation clustering and Set~3 vs Set~4 discriminant maximization.

\paragraph{S4.1 Correlation-based clustering of sub-metrics.}
The cluster partition is fit by hierarchical
agglomerative clustering on the calibrated sub-metric correlation
structure on Set~3 alone. From the calibrated Set~3 scores we compute
the $13 \times 13$ Pearson $|\rho|$ matrix, convert it to a distance
via $d_{ij} = 1 - |\rho_{ij}|$, and apply average-linkage agglomerative
clustering (main-text Figure~\ref{fig:cluster-structure}a). The
dendrogram has its broadest natural gap in merge distance near
$d \approx 0.75$; cutting there returns five clusters, and the choice
is robust within the gap
(Section~\ref{sec:results-cluster-structure}). Within-cluster mean pairwise $|\rho|$
on Set~3 is $0.51$ for C1 (photometric, six sub-metrics), $0.34$ for
C2 (texture / sharpness, three sub-metrics), and $0.38$ for C3 (pose, two sub-metrics); C4 (illumination direction) and
C5 (pitch) are singletons and so have no within-cluster pairwise
$|\rho|$ to report. Set~3 is used for clustering because
the discriminant the cluster weights are subsequently fit against
(Section~\ref{sec:cluster-weights}) is defined on Set~3 vs Set~4
pairs, and Set~3's real within-patient pre/post variation reflects
the clinical co-variation structure the master score is intended to
characterize. As a contrast case, the Set~2 Pearson $|\rho|$
matrix, displayed with the Set~3-derived cluster outlines overlaid
in main-text Figure~\ref{fig:cluster-structure}c, shows the three
pose metrics (pitch, yaw, roll) bundling into a single block on Set~2.
\chg{Set~2's in-plane rotation axis perturbs only roll, so pitch and yaw
are coupled there only through their shared zero-response baseline.
The Set~3 split of pitch (C5) from yaw and roll (C3) is therefore a
property of the clinical data itself.} This procedure yields
the five clusters reported in main-text
Section~\ref{sec:results-cluster-structure}; the partition was inspected
for face-validity against known photographic mechanisms but was not
adjusted to match them.

\paragraph{S4.2 Objective: Set~3 vs Set~4 discriminant maximization.}
Cluster weights are fit by maximizing the master-score
discriminant on the contrast between Set~3 (real within-patient
pre/post pairs, $n = 134$) and Set~4 (cross-publication
mismatched pairs, $n = 134$). For a candidate weight vector
$\mathbf{w}$, the master score on every pair $i$ is the weighted
cluster mean $S_i(\mathbf{w}) = \sum_c w_c\, \bar{s}_{c,i}$, where
$\bar{s}_{c,i}$ is the unweighted mean of cluster~$c$'s calibrated
sub-metric scores on pair~$i$ (Section~\ref{sec:supp-s3}). The
between-set sensitivity index is
\begin{equation}
d'(\mathbf{w}) \;=\; \frac{\mu_{S_3}(\mathbf{w}) - \mu_{S_4}(\mathbf{w})}{\sigma_{\text{pooled}}(\mathbf{w})},
\qquad \sigma_{\text{pooled}}^{2}(\mathbf{w}) = \tfrac{1}{2}\bigl(\sigma_{S_3}^{2}(\mathbf{w}) + \sigma_{S_4}^{2}(\mathbf{w})\bigr),
\label{eq:s4-dprime}
\end{equation}
where $\mu$ and $\sigma$ are the empirical mean and
standard deviation of $S(\mathbf{w})$ on the indicated set. The
cluster weights are chosen to maximize this discriminant,
\begin{equation}
\mathbf{w}^{\star} \;=\; \arg\max_{\mathbf{w}}\; d'(\mathbf{w}),
\label{eq:s4-argmax}
\end{equation}
subject to
\begin{enumerate}
  \item[(i)] $\sum_c w_c = 1$ (probability-simplex constraint);
  \item[(ii)] $w_c \in [w_{\min}, w_{\max}] = [0.02, 0.70]$.
\end{enumerate}
\chg{The floor of $0.02$ prevents any cluster from collapsing to zero,
including clusters whose failure modes the Set~3/Set~4 contrast
under-rewards relative to their clinical relevance. No weight reached
either bound in the reported fit.} The ceiling of
$0.70$ prevents any single cluster from absorbing all the weight,
which would reduce the aggregate to effectively one cluster. \chg{Monotonicity of the master score under single-axis perturbations is
checked afterwards against Set~2 (main-text
Section~\ref{sec:results-weight-fit};
Figure~\ref{fig:dose-response}). It is not a fit constraint, and Set~2
enters the fit nowhere.}

\paragraph{S4.3 Solver, cross-validation, and within-cluster distribution.}
Equation~\ref{eq:s4-argmax} is solved by sequential
quadratic programming~\cite{Kraft1988} from twenty-four
initializations, one uniform and twenty-three drawn from a
symmetric Dirichlet \chg{with concentration $\alpha = 1$ (random seed
$0$)}, and the highest-$d'$ feasible solution is retained. Out-of-fit performance of the resulting weights is
evaluated by 5-fold stratified cross-validation on the Set~3 and
Set~4 samples; the cluster partition is held fixed at the full-Set~3
derivation (Section~S4.1), and only the weights are re-fit per fold.
The cross-validated mean test $d'$ is $2.13 \pm 0.42$
(mean $\pm$ SD across folds), within $0.02$ of the full-data
$d' = 2.15$, indicating that the weight fit does not materially
overfit the Set~3/Set~4 contrast. Within each cluster, the cluster
weight is distributed evenly across members. Fitted values are
reported in main-text Section~\ref{sec:results-weight-fit}.

\paragraph{S4.3.1 Publication-level cross-validation.}\label{sec:supp-s431}
The 5-fold cross-validation reported above splits at the pair level;
pairs from the same publication may therefore appear in both training
and test folds. To rule out within-publication leakage as a source of
the headline discriminant, we additionally performed a 5-fold group
cross-validation that splits at the publication level. The
publication universe is the union of the $46$ Set~3 publications and
the $50$ publications appearing as endpoints of the cross-publication
Set~4 pairs ($51$ publications in total; $45$ shared between Set~3
and Set~4). Publications were partitioned into five folds. A Set~3
pair was assigned to the fold of its publication; a Set~4 pair was
assigned to a fold only if both endpoint publications shared that
fold, and was otherwise dropped from that fold to prevent leakage. \chg{One fold
collapsed to a single
Set~4 test pair after this filter and is excluded; the remaining four
folds were usable.} Within each fold, clusters were frozen to the
full-Set~3 derivation (identical to S4.3) and only the cluster
weights were re-fit via SLSQP on the training publications. \chg{Pooling
across the four usable held-out folds, the publication-level
cross-validated discriminant is} $d' = 2.20$ ($\mathrm{AUC} = 0.936$), within $0.05$
units of the full-data $d' = 2.15$ (95\% CI $[1.83, 2.55]$) ($\mathrm{AUC} = 0.928$, 95\% CI $[0.896, 0.959]$); the
per-fold mean is $d' = 2.27 \pm 0.52$
($\mathrm{AUC} = 0.946 \pm 0.038$). The equal-weight 5-cluster
baseline under the same publication-level splits achieves
$d' = 2.13$ ($\mathrm{AUC} = 0.927$), confirming the Section~S4.4 finding
that the discriminant is supported by the breadth of measurement
rather than by the precise weight vector. Optimism, defined as the
full-data $d'$ minus the pooled held-out $d'$, is $-0.05$; the
headline discriminant therefore does not reflect overfitting to
source publications. \chg{The full per-fold table and weights are available under the code
availability statement in the end matter.}

\paragraph{S4.4 Sensitivity of the master discriminant to aggregation scheme.}
To bound the risk that the headline discriminant
depends on the particular cluster partition derived in Section~S4.1, we
recomputed the master consistency score under five alternative
aggregations and report the resulting Set~3 vs Set~4 discriminant
(Table~\ref{tab:agg-sensitivity}). The published 5-cluster,
$d'$-maximized aggregation attains $d' = 2.15$ (95\% CI $[1.83, 2.55]$) (AUC~$0.928$, 95\% CI $[0.896, 0.959]$). Weighting the same
five clusters equally attains $d' = 2.03$ (AUC~$0.917$), within $0.12$
units of the fitted scheme, evidence that the discriminant is not
materially dependent on the precise weight vector. The earlier shipped-v1
4-cluster sensitivity-equalized aggregation (Set~2-derived,
calibration-optimized rather than discriminant-optimized) attains
$d' = 1.75$ (AUC~$0.892$), and the domain and single-sub-metric
schemes land in the same neighborhood. A reduced 3-metric aggregation
is noticeably worse, which is where breadth of measurement shows: the
discriminant survives a change of weights but not a loss of
measurements.

\begin{table}[h]
\centering
\caption{Sensitivity of the master consistency score's
discriminant performance (Set~3 vs Set~4) to the choice of
aggregation scheme. The published 5-cluster $d'$-maximized scheme
attains $d' = 2.15$ (95\% CI $[1.83, 2.55]$; bootstrap over resampled source publications, $B = 5{,}000$; see main-text Section~\ref{sec:eval}). \chg{Confidence intervals for the alternative aggregation schemes (rows~2--6) were not separately bootstrapped, so the table reports point estimates only.} Weighting the five clusters equally
is within $0.12$ units of the headline, establishing that the headline result is not
materially dependent on the precise weight vector. \chg{The a-priori 4
categories are lighting, color, composition, and quality.} Rows 2--5 are
quoted from the earlier shipped-v1 calibration (Set~2-derived
sensitivity-equalization) and have not been independently re-fit
under v3; row~6 is computed under v3, and its $d'$ sits within $0.12$
units of the headline, suggesting the qualitative ordering of rows~2--5
is preserved.}
\label{tab:agg-sensitivity}
\begin{tabular}{@{}lcc@{}}
\toprule
Aggregation scheme & $d'$ (S3 vs S4) & AUC \\
\midrule
Cluster (5-cluster, this work, $d'$-max) & $+2.15$ & $0.928$ \\
Cluster (4-cluster, shipped-v1 sens.-eq.) & $+1.75$ & $0.892$ \\
Domain (a-priori 4 categories)    & $+1.80$ & $0.899$ \\
Sub-metric (13, no clustering)    & $+1.91$ & $0.912$ \\
$3$-core (reduced 3-metric)       & $+1.39$ & $0.837$ \\
Equal cluster weights (no fitting) & $+2.03$ & $0.917$ \\
\bottomrule
\end{tabular}
\end{table}

\subsection{Endpoint-anchored calibration: sensitivity of the master discriminant}
\label{sec:supp-s6}

\paragraph{S5.1 Motivation.}
The per-sub-metric calibration described in
Methods~\ref{sec:metrics-jnd} anchors a calibrated score of $0.5$ to
the median Set~3 raw absolute difference, an
\emph{empirical-median} anchor. This section quantifies the
sensitivity of the headline master discriminant
($d' = 2.15$, 95\% CI $[1.83, 2.55]$, $\mathrm{AUC} = 0.928$, 95\% CI $[0.896, 0.959]$ on Set~3 vs Set~4) to an
alternative \emph{endpoint-anchored} calibration in which the
upper anchor is the identical-pair contrast (Set~1) and the lower
anchor is the 95th percentile of cross-publication mismatched-pair
differences (Set~4).

\paragraph{S5.2 Anchor definitions.}
Two anchoring schemes are compared:
\begin{description}
\item[Anchor~A (empirical median, published):]
For each sub-metric $m$, the metric-specific monotone transform $T_m$
(square-root, $\ln(1+x)$, or identity; Methods~\ref{sec:metrics-jnd}) is
applied to the raw absolute difference $|\Delta_m|$, and the
transformed value is linearly mapped so that the median of
$T_m(|\Delta_m|)$ over Set~3 is at calibrated score $0.5$.
\item[Anchor~B (endpoint-anchored):]
The same transform $T_m$ is retained, but the linear map is
re-anchored such that $T_m(0) \mapsto 1$ (the identical-pair
ceiling) and the 95th percentile of $T_m(|\Delta_m|)$ over Set~4
$\mapsto 0$ (the mismatched-pair floor). Values are clipped to
$[0, 1]$ outside the anchor range.
\end{description}

\paragraph{S5.3 Recovery of raw differences and validation.}
Set~1 and Set~4 raw differences were not persisted alongside the
published Set~3 raw-difference table. \chg{They were recovered by
inverting the saved per-pair calibrated sub-metric scores through the
Anchor~A mapping;}
per-metric transform shapes were inferred by fitting
$\mathrm{cal} = a + b \cdot T(\mathrm{raw})$ on Set~3 with
$T \in \{\mathrm{identity}, \sqrt{\cdot}, \ln(1{+}x)\}$
($R^2 = 0.82$--$0.995$; $\ln(1+x)$ selected for nine sub-metrics and
$\sqrt{\cdot}$ for four). These inversion-approximation fits are used
only to recover raw differences from the saved calibrated scores and
are distinct from the production calibration transforms of
Table~\ref{tab:s3-calibration}. Inversion fidelity was validated on
Set~1 identical pairs, on which the recovered $|\Delta_m|$ collapsed
to within $10^{-3}$ of zero for all sub-metrics. As a stronger
end-to-end check, Anchor~A was re-applied to the recovered raw
differences and the master discriminant on Set~3 vs Set~4
\chg{recomputed to $d' = 2.153$ and $\mathrm{AUC} = 0.928$, reproducing
the published baseline and confirming that the inversion does not bias
the comparison reported below.}

\paragraph{S5.4 Headline comparison.}
Cluster weights are held at the published v3 fit
(Section~\ref{sec:supp-s4}); only the per-sub-metric calibration
mapping is swapped from Anchor~A to Anchor~B. Headline statistics
on Set~3 versus Set~4 are reported in
Table~\ref{tab:supp-s6-headline}.

\begin{table}[h]
\centering
\caption{Master-score statistics under Anchor~A (median, published)
and Anchor~B (endpoint-anchored). Cluster weights held fixed at the
v3 fit; only the per-sub-metric calibration mapping is swapped.}
\label{tab:supp-s6-headline}
\begin{tabular}{lcc}
\toprule
Statistic                             & Anchor~A & Anchor~B \\
\midrule
$d'$ (Set~3 vs Set~4)                 & $2.153$  & $2.169$  \\
$\mathrm{AUC}$ (empirical)            & $0.928$  & $0.930$  \\
$\mu_{S3}$                            & $0.504$  & $0.505$  \\
$\mu_{S4}$                            & $0.261$  & $0.245$  \\
$\sigma_{\text{pooled}}$              & $0.113$  & $0.120$  \\
Set~1 master mean                     & $0.990$  & $0.995$  \\
Set~4 master 95th percentile          & $0.440$  & $0.432$  \\
\bottomrule
\end{tabular}
\end{table}

\paragraph{S5.5 Per-cluster discriminant.}
The cluster-level $d'$ (Methods~\ref{sec:eval}) is
stable under the anchor swap
(Table~\ref{tab:supp-s6-percluster}). \chg{All five clusters' marginal
discriminants shift by $|\Delta d'| \leq 0.05$ and all remain positive.
The two strongest exchange rank: C1 and C2 read $1.39$ and $1.41$
under Anchor~A, and $1.44$ and $1.42$ under Anchor~B.}

\begin{table}[h]
\centering
\caption{Per-cluster $d'$ on Set~3 vs Set~4 under each anchor.}
\label{tab:supp-s6-percluster}
\begin{tabular}{lcc}
\toprule
Cluster                               & Anchor~A & Anchor~B \\
\midrule
C1 Photometric                        & $1.39$   & $1.44$   \\
C2 Texture / sharpness                & $1.41$   & $1.42$   \\
C3 Pose                               & $0.72$   & $0.72$   \\
C4 Illumination direction             & $0.94$   & $0.95$   \\
C5 Pitch                              & $1.03$   & $1.03$   \\
\bottomrule
\end{tabular}
\end{table}

\paragraph{S5.6 Interpretation.}
\chg{Re-anchoring moves the master $d'$ by $+0.016$ and the empirical
AUC by $+0.002$. The separation therefore does not depend on the
empirical-median anchor. Anchor~B raises the Set~1 ceiling and
compresses the mismatched-pair distribution, both by construction.}

\subsection{Provenance of the sub-metrics}
\label{sec:supp-provenance}

Table~\ref{tab:provenance} maps each published capture requirement to the
sub-metric that measures it, and lists the requirements for which no
sub-metric exists.

\begin{table}[H]
  \centering
  \footnotesize
  \caption{Capture requirements from published photographic standards and the
  sub-metrics that measure them. The middle block lists sub-metrics included
  on image-quality grounds, for which no photographic standard specifies a
  requirement. \chg{Requirements in the final block have no corresponding sub-metric
  here. Focal length and subject-to-camera distance cannot be recovered
  from an image pair without capture metadata; the remaining three were
  not implemented.}}
  \label{tab:provenance}
  \begin{tabular}{@{}p{42mm}p{52mm}p{40mm}@{}}
    \toprule
    \textbf{Requirement} & \textbf{Sub-metric(s)} & \textbf{Source} \\
    \midrule
    Same lighting / illumination; appropriate exposure & Brightness,
      contrast, lightness offset & \cite{DiBernardo1998, Wolfe2020} \\
    \addlinespace
    No cast shadow; lighting angle & Shadow extent, illumination direction &
      \cite{Wolfe2020} \\
    \addlinespace
    Same film / sensor colour response and illumination colour &
      Global $\Delta E_{00}$, median-pixel $\Delta E_{00}$ &
      \cite{DiBernardo1998, Wolfe2020} \\
    \addlinespace
    Adequate focus & Image gradient, focus &
      \cite{DiBernardo1998, Wolfe2020} \\
    \addlinespace
    Same patient position; consistent view & Yaw, pitch, roll &
      \cite{DiBernardo1998, Wolfe2020} \\
    \midrule
    \emph{No standard requirement} & Noise & \cite{Wang2004,DonohoJohnstone1994} \\
    \midrule
    Same lens focal length and lens setting & --- & \cite{DiBernardo1998} \\
    Same subject-to-camera distance & --- & \cite{DiBernardo1998} \\
    Uniform background & --- & \cite{Wolfe2020} \\
    Preparation (clothing, jewellery, makeup) & --- & \cite{Wolfe2020} \\
    Facial expression & --- & \cite{Wolfe2020} \\
    \bottomrule
  \end{tabular}
\end{table}

\end{document}